%% file: main.tex
\documentclass{article}

\usepackage[margin = 0.9in]{geometry}
\makeatletter
\renewcommand\normalsize{%
  \@setfontsize\normalsize\@xpt{11}%
  \abovedisplayskip 10\p@ \@plus2\p@ \@minus5\p@
  \abovedisplayshortskip \z@ \@plus3\p@
  \belowdisplayshortskip 6\p@ \@plus3\p@ \@minus3\p@
  \belowdisplayskip \abovedisplayskip
  \let\@listi\@listI}
\normalsize
\makeatother
\usepackage{parskip}
\usepackage[compress, sort, round]{natbib}
\usepackage[x11names,table,xcdraw]{xcolor}

\usepackage{amsthm,amssymb,mathtools}
\newtheorem{proposition}{Proposition}

\input{math_commands.tex}
\input{packages.tex}

\usepackage[labelfont=bf]{caption}
\usepackage{flafter}
\usepackage{fontawesome5}


\title{\bfseries Integral DeltaNet: Improved Linear Attention \\ with Adaptive Memory Integration}

\author{
Junsoo Oh\textsuperscript{1}\thanks{This work was done during an internship at KRAFTON AI.} \quad
Jihun Yun\textsuperscript{2} \quad
Junhyuck Kim\textsuperscript{2} \quad
Hanbaek Lyu\textsuperscript{3} \\[2pt]
\normalsize
\begin{tabular}[t]{@{}c@{}}
\textsuperscript{1}Kim Jaechul Graduate School of AI, KAIST \qquad \textsuperscript{2}KRAFTON AI \\
\textsuperscript{3}Department of Mathematics, University of Wisconsin--Madison
\end{tabular}
}

\date{\normalsize\faGithub~\href{https://github.com/junsoo424/Integral-DeltaNet}{\texttt{github.com/junsoo424/Integral-DeltaNet}}}

\begin{document}
\pagenumbering{arabic}
\maketitle

\begin{abstract}
Linear attention replaces softmax attention with a linear formulation, enabling linear-time computation with a fixed-size recurrent state instead of a growing key--value cache.
These computational and memory advantages have attracted growing interest in linear attention for long-context language modeling.
However, compressing past information into a fixed-size state causes \emph{memory collisions} that hinder long-context understanding. 
Although DeltaNet mitigates memory collisions through delta-rule updates and Gated DeltaNet extends this approach with gating, these mechanisms may also discard information relevant to future predictions.
To address this limitation, we introduce \emph{Integral DeltaNet}, which maintains two complementary recurrent states: a fast-updating \emph{delta state} that captures current key--value associations and an \emph{integral state} that preserves a weighted history of the delta state.
At readout, the model adaptively combines these states to access information across different timescales.
We further derive a chunkwise parallel formulation that enables hardware-efficient training.
We evaluate Integral DeltaNet using 340M and 1.3B parameter models trained from scratch on 15B and 100B FineWeb-Edu tokens, respectively. Comparisons with strong recurrent baselines show improved overall performance in language modeling, commonsense reasoning, real-world and synthetic retrieval, long-context understanding, and length extrapolation. We also compare training throughput and activation memory, as well as decoding latency and memory.
\end{abstract}

\input{Sections/01Introduction}
\input{Sections/02Preliminaries}

\input{Sections/03Method}
\input{Sections/04Experiments}

\input{Sections/05RelatedWork}

\input{Sections/06Conclusion}

\subsection*{Acknowledgement}
 This research was supported by a grant from KRAFTON AI.

\bibliography{Reference}
\bibliographystyle{plainnat}

\clearpage
\appendix

\input{Sections/AProof}
\input{Sections/BGateAblation}

\end{document}

%% file: math_commands.tex
\usepackage{amsmath,amsfonts,bm}

\newcommand{\diag}{\mathrm{Diag}}
\def\valpha{{\bm{\alpha}}}
\def\vgamma{{\bm{\gamma}}}
\newcommand{\vlambda}{\bm{\lambda}}
\def\eqref#1{(\ref{#1})}
\def\vone{{\bm{1}}}
\def\vb{{\bm{b}}}
\def\vh{{\bm{h}}}
\def\vk{{\bm{k}}}
\def\vo{{\bm{o}}}
\def\vq{{\bm{q}}}
\def\vv{{\bm{v}}}
\def\vw{{\bm{w}}}
\def\vz{{\bm{z}}}
\def\mH{{\bm{H}}}
\def\mI{{\bm{I}}}
\def\mK{{\bm{K}}}
\def\mM{{\bm{M}}}
\def\mO{{\bm{O}}}
\def\mP{{\bm{P}}}
\def\mQ{{\bm{Q}}}
\def\mS{{\bm{S}}}
\def\mV{{\bm{V}}}
\def\mW{{\bm{W}}}
\def\mX{{\bm{X}}}
\def\mZ{{\bm{Z}}}
\newcommand{\R}{\mathbb{R}}
\def\mzero{{\bm{0}}}
\def\vomega{{\bm{\omega}}}

%% file: packages.tex
\usepackage[utf8]{inputenc}
\usepackage[T1]{fontenc}
\usepackage{hyperref}
\usepackage{url}
\usepackage{graphicx}
\usepackage{booktabs}
\usepackage{colortbl}
\definecolor{kdashade}{HTML}{FBECF2}
\definecolor{gdn2shade}{HTML}{E6F1E6}
\newcommand{\kdarow}{\rowcolor{kdashade}}
\newcommand{\gdntworow}{\rowcolor{gdn2shade}}
\newcommand{\best}[1]{#1$^{\star}$}
\newcommand{\second}[1]{#1$^{\dagger}$}
\usepackage{microtype}
\usepackage{float}

\usepackage{pgfplots}
\pgfplotsset{compat=1.18}
\usepgfplotslibrary{groupplots}

\pgfplotsset{
  paperplot/.style={
    font=\rmfamily\fontsize{10}{12}\selectfont,
    title style={font=\rmfamily\fontsize{10}{12}\selectfont},
    label style={font=\rmfamily\fontsize{7}{8.4}\selectfont},
    tick label style={font=\rmfamily\fontsize{8}{9.6}\selectfont},
    legend style={font=\rmfamily\fontsize{7}{8.4}\selectfont},
    axis line style={draw=gray!55, line width=0.75pt},
    major tick style={line width=0.6pt},
    minor tick style={line width=0.4pt},
    tick align=inside,
    every axis plot/.append style={line width=1pt, mark size=1.5pt},
  },
}

\definecolor{darkblue}{rgb}{0.0,0.0,0.65}
\definecolor{darkred}{rgb}{0.65,0.0,0.0}
\hypersetup{
	colorlinks = true,
	citecolor  = darkblue,
	linkcolor  = darkred,
	filecolor  = darkblue,
	urlcolor   = darkblue,
}

%% file: Sections/01Introduction.tex
\section{Introduction}\label{sec:intro}
Most modern large language models (LLMs) are built on the Transformer architecture~\citep{vaswani2017attention}, whose softmax attention enables effective sequence modeling and hardware-efficient parallel training on GPUs.
However, its computation scales quadratically with context length, and the key--value cache grows linearly during autoregressive inference.
The demand for longer contexts has motivated alternatives with linear computational complexity and constant memory usage in sequence length.
Linear attention~\citep{katharopoulos2020transformers} achieves this through a kernelized dot-product formulation that admits a matrix-valued linear recurrence, supporting parallel training and efficient recurrent inference.
Although early linear attention models~\citep{choromanski2021rethinking,peng2021random} struggled to match Transformers in language modeling, recent variants have become increasingly competitive and have been incorporated into frontier-scale models such as Qwen3-Next~\citep{qwen2025next} and Kimi K3~\citep{team2026kimi}.

\input{Sections/fig_integral_framework}

Recent linear attention architectures control memory through decay or gating, as in RetNet~\citep{sun2023retentive}, Mamba-2~\citep{dao2024transformers}, and Gated Linear Attention~\citep{yang2024gated}, or delta-rule updates, as in DeltaNet~\citep{schlag2021linear,yang2024parallelizing}.
These mechanisms mitigate \emph{memory collisions} caused by compressing many key--value associations into a fixed-size state~\citep{schlag2021linear}.
Gating reduces previously stored associations, whereas the delta rule performs targeted updates by removing the value currently associated with an incoming key before writing the new association.
Gated DeltaNet (GDN)~\citep{yang2025gated} combines these mechanisms, while Kimi Delta Attention (KDA)~\citep{team2025kimi} and Gated DeltaNet-2 (GDN-2)~\citep{hatamizadeh2026gated} introduce more expressive updates.
However, both decay and targeted overwriting can discard information needed over long contexts. In this work, we address the following question: 

\begin{center}
    \emph{Can we mitigate memory collisions while better retaining information needed for the future?} 
\end{center}
We propose \emph{Integral DeltaNet}, which addresses this limitation using two states operating at different timescales: a fast \emph{delta state} $S_t$, updated using an expressive linear-attention rule such as Kimi Delta Attention or Gated DeltaNet-2, and a slow \emph{integral state} $X_t$, which maintains a channel-wise weighted history of past delta states. As illustrated in Figure~\ref{fig:integral-framework}, the model adaptively combines the two states at readout, enabling it to access information across multiple timescales.

We derive a chunkwise parallel formulation for Integral DeltaNet that enables hardware-efficient training and use it to train 340M- and 1.3B-parameter models from scratch on 15B and 100B tokens, respectively.
We compare these models with strong recurrent baselines and Transformer++ on language modeling, commonsense reasoning, and real-world retrieval at both scales.
We further evaluate the 1.3B models on tasks requiring longer contexts: synthetic retrieval, long-context understanding, and length extrapolation.
We observe gains over the backbones across multiple tasks and also compare training throughput and activation memory, as well as decoding latency and memory.

%% file: Sections/fig_integral_framework.tex
\suppressfloats[t]
\begin{figure}[t]
\centering
\begingroup
\definecolor{idnink}{HTML}{283440}
\definecolor{idnline}{HTML}{64727F}
\definecolor{idnblue}{HTML}{356B96}
\definecolor{idnbluebg}{HTML}{EAF2F9}
\definecolor{idnamber}{HTML}{996A24}
\definecolor{idnamberbg}{HTML}{FCF2DE}
\definecolor{idngraybg}{HTML}{F5F7F9}
\resizebox{0.85\linewidth}{!}{%
\begin{tikzpicture}[
x=1cm,y=.78cm,font=\sffamily\fontsize{9}{11}\selectfont,text=idnink,
state/.style={rounded corners=3pt,line width=.8pt,align=center,inner xsep=5pt,inner ysep=4pt,text width=2.65cm,minimum height=1.15cm},
flow/.style={->,>=stealth,line width=.85pt},
trace/.style={dashed,line width=.8pt},
note/.style={font=\sffamily\fontsize{8}{10}\selectfont,align=center}]
\path[use as bounding box] (0,-.65) rectangle (14,5.75);
\node[anchor=west,font=\sffamily\bfseries\fontsize{10}{12}\selectfont] at (.25,5.55) {Context};
\node[note,anchor=west,text=idnline] at (2.1,5.55) {Earlier};
\draw[flow,draw=idnline!65] (3.65,5.55) -- (11.1,5.55);
\node[note,anchor=east,text=idnline] at (13.5,5.55) {Recent / current};
\foreach \i in {0,...,19} {
 \draw[draw=idnline!55,fill=idngraybg,line width=.55pt,rounded corners=.5pt]
   ({.65+.65*\i-.22},4.78) rectangle ({.65+.65*\i+.22},5.22);
}
\node[state,draw=idnamber,fill=idnamberbg] (x) at (3.8,2.6)
 {\textbf{Integral state}\\$\mX_t$\\Retained history};
\node[state,draw=idnblue,fill=idnbluebg] (s) at (10.3,2.6)
 {\textbf{Delta state}\\$\mS_t$\\KDA / GDN-2};
\draw[trace,draw=idnamber] (1.95,4.76) -- (x.north);
\draw[trace,draw=idnamber] (5.2,4.76) -- (x.north);
\draw[trace,draw=idnamber] (8.45,4.76) -- (x.north);
\draw[trace,draw=idnblue] (9.1,4.76) -- (s.north);
\draw[trace,draw=idnblue] (11.05,4.76) -- (s.north);
\draw[trace,draw=idnblue] (13,4.76) -- (s.north);
\node[note,text=idnamber,fill=white,inner sep=3pt] at (3.8,3.88) {Memory traces};
\node[note,text=idnblue,fill=white,inner sep=3pt] at (10.3,3.88) {Current / recent associations};
\node (xprev) at (.65,2.6) {$\mX_{t-1}$};
\node (sprev) at (13.5,2.6) {$\mS_{t-1}$};
\draw[flow,draw=idnamber] (xprev.east) -- node[above,note] {$\vone-\vlambda_t$} (x.west);
\draw[flow,draw=idnblue] (sprev.west) -- (s.east);
\draw[flow,draw=idnamber] (s.west) -- node[above,note,text=idnamber] {Integrate with $\vlambda_t$} (x.east);
\node[circle,draw=idnink,fill=white,line width=.8pt,minimum size=.55cm,inner sep=0pt,font=\fontsize{12}{14}\selectfont] (sum) at (7.05,.65) {$+$};
\draw[flow,draw=idnamber] (x.south) -- (3.8,.65) -- (sum.west);
\draw[flow,draw=idnblue] (s.south) -- (10.3,.65) -- (sum.east);
\node[note,text=idnamber] at (5.2,1.08) {$\mX_t((\vone-\vgamma_t)\odot\vq_t)$};
\node[note,text=idnblue] at (8.85,1.08) {$\mS_t(\vgamma_t\odot\vq_t)$};
\node[align=center] (out) at (7.05,-.3) {\textbf{Gated readout} $\vo_t$};
\draw[flow,draw=idnink] (sum.south) -- (out.north);
\draw[trace,draw=idnline] (.25,.95) -- (.85,.95);
\node[note,anchor=west,align=left,text=idnline,inner sep=2pt] at (.95,.95) {Context retained\\(schematic)};
\draw[flow,draw=idnline] (.25,.05) -- (.85,.05);
\node[note,anchor=west,align=left,text=idnline,inner sep=2pt] at (.95,.05) {State computation};
\end{tikzpicture}%
}
\endgroup
\caption{\textbf{Integral DeltaNet.} The integral state accumulates a weighted history of delta states. At readout, the model adaptively combines both states to access information across timescales.}
\label{fig:integral-framework}
\end{figure}
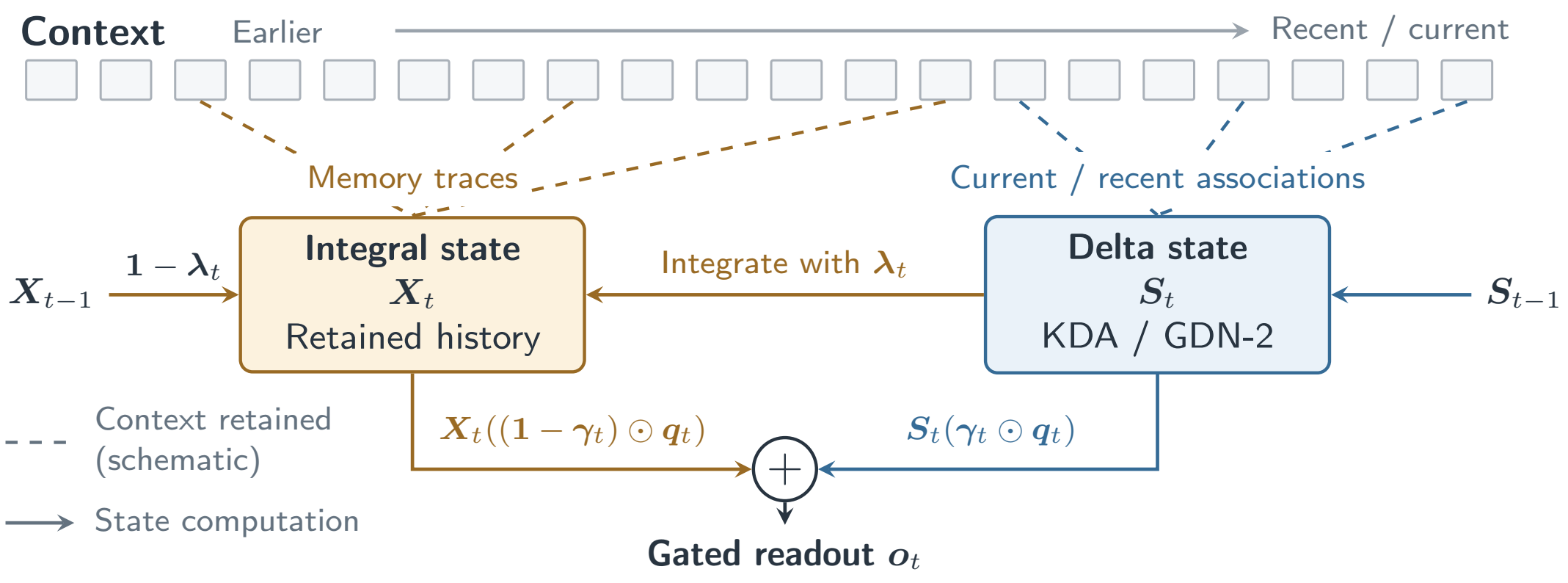

%% file: Sections/02Preliminaries.tex
\section{Preliminaries}\label{sec:background}
\subsection{Linear Attention}
Linear attention~\citep{katharopoulos2020transformers} replaces softmax-normalized query--key similarities with a kernelized dot product, yielding the output at position $t$ as
\begin{equation*}
    \vo_t
    = \sum_{i=1}^t \vv_i (\vk_i^\top \vq_t)
    = \left( \sum_{i=1}^t \vv_i \vk_i^\top \right) \vq_t
    \in \R^{d_v},
\end{equation*}
where $\vk_t, \vq_t \in \R^{d_k}$ and $\vv_t \in \R^{d_v}$ denote the key, query, and value vectors, respectively. By maintaining a prefix state $\mS_t$, the same computation can be written in recurrent form as
\begin{equation}\label{eq:linear-attention-update}
    \mS_0 = \mathbf{0}, \qquad
    \mS_t = \mS_{t-1} + \vv_t \vk_t^\top \in \R^{d_v \times d_k},
    \qquad
    \vo_t = \mS_t \vq_t \in \R^{d_v}.
\end{equation}
For a sequence of length $L$, the equivalent masked matrix form is
\begin{equation*}
    \mO = \left( \mQ \mK^\top \odot \mM \right) \mV
    \in \R^{L \times d_v},
\end{equation*}
where $\mQ, \mK \in \R^{L \times d_k}$ and $\mV, \mO \in \R^{L \times d_v}$ are the query, key, value, and output matrices, respectively. The causal mask $\mM \in \{0,1\}^{L \times L}$ satisfies $M_{t,i}=1$ for $i \leq t$ and $M_{t,i}=0$ otherwise.

\paragraph{Chunkwise Parallel Training.}
The recurrent form scales linearly with the sequence length $L$ but processes tokens sequentially, whereas the fully parallel form exploits efficient matrix multiplications at the cost of materializing an $L\times L$ interaction matrix. Chunkwise training~\citep{hua2022transformer,sun2023retentive,yang2024parallelizing} bridges this gap by partitioning the sequence into chunks of size $C$, evaluating token interactions in parallel within each chunk, and propagating only a fixed-size state across chunks. Let $N=\lceil L/C\rceil$ be the number of chunks. For chunk $[t]$, where $t\in\{0,\ldots,N-1\}$, let $\mQ_{[t]},\mK_{[t]}\in\R^{C\times d_k}$ and $\mV_{[t]},\mO_{[t]}\in\R^{C\times d_v}$ denote the row-stacked query, key, value, and output matrices, and let $\mS_{[t]}:=\mS_{tC}$ denote the incoming state. Then, chunkwise form can be written as
\begin{equation*}
    \mO_{[t]}
    = \mQ_{[t]}\mS_{[t]}^\top
    + \left(\mQ_{[t]}\mK_{[t]}^\top\odot\mM_C\right)\mV_{[t]},
    \qquad
    \mS_{[t+1]}
    = \mS_{[t]} + \mV_{[t]}^\top\mK_{[t]},
\end{equation*}
where $\mM_C\in\{0,1\}^{C\times C}$ is the causal mask within each chunk. This reduces the sequential depth from $L$ to approximately $L/C$ without constructing a global attention matrix. The same principle extends to many linear attention variants and underlies FlashLinearAttention~\citep{yang2024fla}, whose fused chunkwise kernels enable efficient training on modern accelerators.

\subsection{Memory Collision}Linear attention compresses all past key--value associations into the fixed-size state $\mS_t$. Assuming unit-normalized keys, retrieving the value associated with $\vk_j$ gives
\begin{equation*}    
    \mS_t\vk_j    = \vv_j    + \sum_{i=1,  i\neq j}^t      (\vk_i^\top\vk_j)\vv_i.
\end{equation*}The second term captures interference from other associations and vanishes when all other keys are orthogonal to $\vk_j$; otherwise, it is generally nonzero. Therefore, storing a growing number of associations in the same state inevitably causes \emph{memory collision}, particularly over long contexts.

DeltaNet mitigates this interference using a delta-rule update with input-dependent strength $\beta_t=\operatorname{sigmoid}(\vw_\beta^\top\vh_t)$, where $\vh_t$ is the input hidden state. This gate controls both erasure of the previously stored value and writing of the new key--value association:
\begin{equation}\label{eq:deltanet-update}
    \mS_t
    =
    \mS_{t-1}\big(\mI-\underbrace{\beta_t\vk_t\vk_t^\top}_{\text{erase old}}\big)
    +
    \underbrace{\beta_t\vv_t\vk_t^\top}_{\text{write new}}
    =
    \mS_{t-1}
    +
    \beta_t\left(\vv_t-\mS_{t-1}\vk_t\right)\vk_t^\top.
\end{equation}

Mamba-2 takes a complementary approach, using a data-dependent decay to suppress old information, parameterized as $\alpha_t = \exp\bigl(-\operatorname{softplus}(\vw_\Delta^\top\vh_t+b_\Delta) \exp(a)\bigr)$ with a learnable scalar $a$ per head:
\begin{equation}\label{eq:mamba2-update}
    \mS_t = \alpha_t \mS_{t-1} + \vv_t\vk_t^\top.
\end{equation}
Gated DeltaNet combines these complementary approaches, coupling data-dependent decay with the targeted overwriting of the delta rule:
\begin{equation}\label{eq:event-gdn}
    \mS_t
    = \alpha_t\mS_{t-1}\left(\mI-\beta_t\vk_t\vk_t^\top\right)
    + \beta_t\vv_t\vk_t^\top.
\end{equation}
Kimi Delta Attention and Gated DeltaNet-2 further increase the expressivity of this framework through channel-wise update mechanisms: Kimi Delta Attention replaces scalar decay with per-channel gating, and Gated DeltaNet-2 introduces separate channel-wise erase and write gates.

These mechanisms reduce memory collisions, but information removed by decay or overwriting may later become important and cannot be recovered from the recurrent state. In this work, we address this trade-off between reducing interference and preserving potentially useful memory.

\begin{figure}[t]
\centering\color{black}
\begin{tikzpicture}[
  x=0.0033\linewidth,y=0.0033\linewidth,
  every node/.style={font=\footnotesize,inner sep=1pt},
  good/.style={rounded corners=1.5pt,draw=green!40!black,fill=green!9,minimum width=16pt,minimum height=14pt,text=green!30!black},
  bad/.style={rounded corners=1.5pt,draw=red!60!black,fill=red!8,minimum width=16pt,minimum height=14pt,text=red!65!black},
  truth/.style={rounded corners=1.5pt,draw=black!35,fill=black!3,minimum width=16pt,minimum height=14pt}
]
\path[use as bounding box] (0,20) rectangle (222,101);
\node[font=\footnotesize\bfseries] at (23,95) {Unit keys};
\node[font=\footnotesize\bfseries] at (77,95) {Values};
\draw[black!30,dashed,line width=.45pt] (4.813,69.5)--(23,80)--(41.187,69.5);
\draw[->,violet,line width=.9pt] (23,59)--(23,80);
\draw[->,blue!65!black,line width=.9pt] (23,59)--(4.813,69.5);
\draw[->,blue!65!black,line width=.9pt] (23,59)--(41.187,69.5);
\fill[black!70] (23,59) circle (.8pt);
\node[text=violet] at (23,86) {$\vk_1$};
\node[text=blue!65!black] at (5,77) {$\vk_2$};
\node[text=blue!65!black] at (42,77) {$\vk_3$};
\node at (23,49) {$\vk_1=\vk_2+\vk_3$};
\node at (23,38) {$\|\vk_t\|_2=1$};
\draw[black!40,line width=.4pt] (65,63)--(69,63)--(69,59);
\draw[->,violet,line width=.9pt] (65,59)--(65,84.456);
\draw[->,blue!65!black,line width=.9pt] (65,59)--(83,59);
\fill[black!70] (65,59) circle (.8pt);
\node[anchor=west,text=violet] at (68,81) {$\vv_{\mathrm N}$};
\node[anchor=west,text=blue!65!black] at (85,59) {$\vv_{\mathrm Y}$};
\node[text=violet] at (77,49) {$\|\vv_{\mathrm N}\|_2^2=2$};
\node[text=blue!65!black] at (77,38) {$\|\vv_{\mathrm Y}\|_2^2=1$};
\node at (50,26) {$(\vv_1,\vv_2,\vv_3)=(\vv_{\mathrm N},\vv_{\mathrm Y},\vv_{\mathrm Y})$};
\draw[black!25,line width=.4pt] (110,24)--(110,98);
\begin{scope}[shift={(122,80)}]
\node[anchor=west] at (0,9) {Query};
\node at (49,9) {$\vk_1$};
\node at (65,9) {$\vk_2$};
\node at (81,9) {$\vk_3$};
\node at (97,9) {$\mathcal R$};
\node[anchor=west] at (0,-4) {Correct};
\node[truth] at (49,-4) {N};
\node[truth] at (65,-4) {Y};
\node[truth] at (81,-4) {Y};
\node[anchor=west,font=\scriptsize] at (0,-20) {Linear Attention};
\node[good] at (49,-20) {N};
\node[bad] at (65,-20) {N};
\node[bad] at (81,-20) {N};
\node at (97,-20) {$\frac23$};
\node[anchor=west,align=left,font=\scriptsize] at (0,-39) {Mamba-2 /\\DeltaNet /\\Gated DeltaNet};
\node[bad] at (49,-39) {Y};
\node[good] at (65,-39) {Y};
\node[good] at (81,-39) {Y};
\node at (97,-39) {$\frac13$};
\end{scope}
\end{tikzpicture}
\caption{Illustrative example and retrieval performance.}
\label{fig:illustrative-example}
\end{figure}
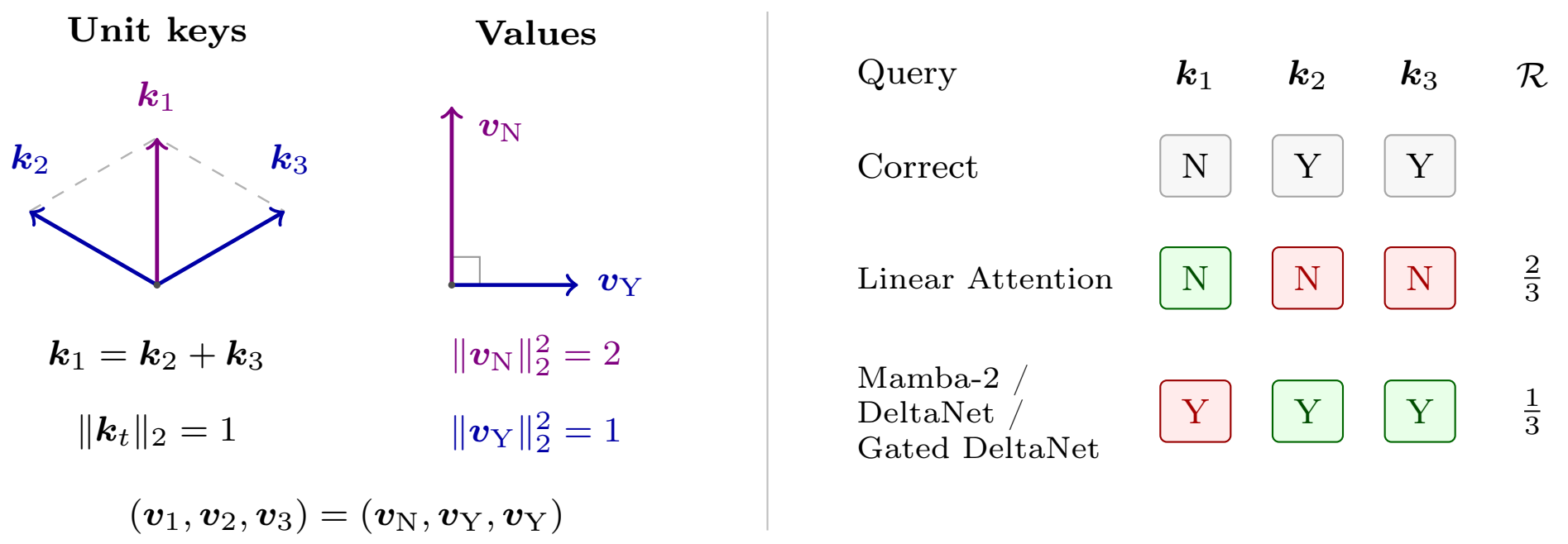

\paragraph{Illustrative Example.}
To illustrate this trade-off (Figure~\ref{fig:illustrative-example}), we consider three sequential key--value pairs $(\vk_t,\vv_t)_{t=1}^3$ with keys $\vk_t\in\R^2$. Let $\vv_{\mathrm Y},\vv_{\mathrm N}\in\R^{d_v}$, $d_v\ge2$, represent \textsf{Yes} and \textsf{No}, respectively. We choose the keys and values to satisfy
\begin{equation*}
\begin{gathered}
 \|\vk_t\|_2=1,\qquad \vk_1=\vk_2+\vk_3,\\
 \vv_{\mathrm Y} \perp \vv_{\mathrm N},\quad
 \|\vv_{\mathrm Y}\|_2^2=1,\quad \|\vv_{\mathrm N}\|_2^2=2,\\
 (\vv_1,\vv_2,\vv_3)=(\vv_{\mathrm N},\vv_{\mathrm Y},\vv_{\mathrm Y}).
\end{gathered}
\end{equation*}

We compare the state updates for linear attention~\eqref{eq:linear-attention-update}, Mamba-2~\eqref{eq:mamba2-update}, DeltaNet~\eqref{eq:deltanet-update}, and Gated DeltaNet~\eqref{eq:event-gdn}. Each $\alpha_t, \beta_t$ may vary across writes within $(0,1]$. Given the final memory state $\mS_3$, we sample a query uniformly from the stored keys:
\[
 \vq_3\sim\mathrm{Unif}\{\vk_1,\vk_2,\vk_3\}.
\]

Each model $\mathcal M$ (Linear Attention, Mamba-2, DeltaNet, Gated DeltaNet) reads $\mS_3\vq_3$ and predicts $\widehat y_3^{\mathcal M}=\arg\max_{y\in\{\mathrm Y,\mathrm N\}}\vv_y^\top\mS_3\vq_3$. The correct answers are $y(\vk_1)=\mathrm N$ and $y(\vk_2)=y(\vk_3)=\mathrm Y$. We define the retrieval risk as $\mathcal R_{\mathcal M}=\Pr_{\vq_3}[\widehat y_3^{\mathcal M}\ne y(\vq_3)]$, where the probability is over the uniformly sampled query.

Below in Proposition~\ref{prop:event-gdn}, we show that there exists a trade-off between memory collision and forgetting. 
\begin{proposition}\label{prop:event-gdn}
Under the setup above, the retrieval risks satisfy
\begin{equation}\label{eq:event-gdn-error}
 \mathcal R_{\mathrm{Linear Attention}}=\frac23, \qquad
 \min_{\{\alpha_t\}}\mathcal R_{\mathrm{Mamba\textup{-}2}}
 =\min_{\{\beta_t\}}\mathcal R_{\mathrm{DeltaNet}}
 =\min_{\{\alpha_t,\beta_t\}}\mathcal R_{\mathrm{Gated DeltaNet}}
 =\frac13.
\end{equation}
\end{proposition}
In Linear Attention, the initial \textsf{No} association dominates the readouts for both later \textsf{Yes} keys, causing two retrieval errors. Mamba-2 and DeltaNet can reduce this interference through decay and targeted overwriting, respectively, and recover both later \textsf{Yes} answers. However, both models forget the first \textsf{No} answer, illustrating how suppressing interference can weaken earlier information still needed for retrieval.

Gated DeltaNet combines decay and overwriting, but it still cannot recover both later \textsf{Yes} answers without forgetting the first \textsf{No} answer. This limitation is inherent to the structure of our example: since $\vk_1=\vk_2+\vk_3$, any single-state linear readout satisfies $\mS_3\vk_1=\mS_3\vk_2+\mS_3\vk_3$. Hence, two readouts favoring \textsf{Yes} also make the first favor \textsf{Yes}. Thus, even with optimal gates, Gated DeltaNet cannot achieve zero retrieval error in this example; its minimum risk remains $1/3$. Although this limitation follows from the chosen key structure, the example provides a concrete illustration of the trade-off discussed above: the same updates that reduce memory collisions can also cause forgetting of earlier associations that remain relevant to subsequent queries. The full proof is in Appendix~\ref{app:event-gdn}.

%% file: Sections/03Method.tex
\section{Integral DeltaNet}\label{sec:method}

In this work, we propose \emph{Integral DeltaNet} to reduce memory interference while preserving useful information from distant past. Figure~\ref{fig:integral-framework} illustrates the central idea: maintain an adaptive \emph{delta state} for the current context alongside an \emph{integral state} that retains a weighted history of earlier delta states. Reading both states gives the model access to information that subsequent updates may have weakened. The two-state recurrence also supports hardware-efficient chunkwise parallel training.

\subsection{Delta States and Integral States}\label{subsec:state}
Our Integral DeltaNet consists of a fast delta state $\mS_t$ and a slower integral state $\mX_t$, defined below. We update $\mS_t$ using the expressive channel-wise rules of Kimi Delta Attention and Gated DeltaNet-2:
\begin{align*}
    \text{KDA:} \quad \mS_t &= \mS_{t-1} \diag \left(\valpha_t\right) \left( \mI - \beta_t \vk_t \vk_t^\top \right) +  \beta_t \vv_t \vk_t^\top   \\
    \text{GDN-2:} \quad \mS_t &= \mS_{t-1} \diag \left(\valpha_t\right) \left( \mI -  \left( \vb_t \odot \vk_t \right) \vk_t^\top \right) + \left(\vw_t \odot \vv_t \right) \vk_t^\top
\end{align*}

We define the \emph{integral state} $\mX_t$ by
\begin{equation}\label{eq:integral_state}
    \mX_t = \mS_t \diag \left(\vlambda_t\right) + \mX_{t-1} \diag \left(\vone - \vlambda_t\right),
\end{equation}
where $\mX_0=\mzero$ and $\vlambda_t\in[0,1]^{d_k}$ specifies the integration rate in each key channel. Given a query $\vq_t$ and a channel-wise readout gate $\vgamma_t \in [0,1]^{d_k}$, we compute the output as
\begin{equation}\label{eq:readout-gate}
    \vo_t = \mS_t \left(\vgamma_t \odot \vq_t\right) + \mX_t \left(\left(\vone-\vgamma_t\right)\odot \vq_t\right).
\end{equation}

Figure~\ref{fig:integral-framework} illustrates how the two states complement each other. The delta state is updated with incoming key--value pairs, while the integral state accumulates a weighted history of delta states, preserving contributions that subsequent updates may weaken. Smaller integration rates retain this history for longer, while larger rates make $\mX_t$ follow $\mS_t$ more closely. At readout, larger entries of $\vgamma_t$ place more weight on the current delta state in the corresponding key channels, favoring current information over accumulated history. These gates allow the model to balance information from different timescales according to the current context. Note that the two states are combined only at readout, leaving the delta-state updates unchanged.

\paragraph{Benefits of the Integral State.}We revisit the illustrative example analyzed in Proposition~\ref{prop:event-gdn} to show how Integral DeltaNet improves retrieval performance. We use the same key--value pairs and query distribution, and retain the Gated DeltaNet update~\eqref{eq:event-gdn} for the delta state. For simplicity, we use scalar gates for integration and readout:
\begin{equation}\label{eq:simple_integral}
 \mX_t=\lambda_t\mS_t+(1-\lambda_t)\mX_{t-1},
 \qquad
 \vo_3=\bigl[\gamma_3\mS_3+(1-\gamma_3)\mX_3\bigr]\vq_3,
\end{equation}
where $\mX_0=\mzero$, $\lambda_t\in(0,1)$ is the integration rate at step $t$, and $\gamma_3\in[0,1]$ can depend on the query $\vq_3$. The model selects the answer maximizing $\vv_y^\top\vo_3$. Its retrieval risk is $\mathcal R_{\mathrm{Integral}}$.

The integral state can preserve the first answer even when the delta state loses it, allowing adaptive readout to recover all three answers.
\begin{proposition}\label{prop:event-integral}
    In the illustrative example shown in Figure~\ref{fig:illustrative-example},
Integral DeltaNet~\eqref{eq:simple_integral} can achieve zero retrieval risk, i.e.,
$\min_{\{\alpha_t,\beta_t,\lambda_t\},\,\gamma_3(\cdot)}
\mathcal R_{\mathrm{Integral}}=0$.
\end{proposition}

One construction uses $\beta_t=1$, $(\alpha_1,\alpha_2,\alpha_3)=(1,1/4,1)$, and $\lambda_t=1/8$ for $t=1,2,3$. With these updates, the delta state $\mS_3$ suppresses interference from the initial \textsf{No} association enough to recover both later \textsf{Yes} answers, but loses the first answer. However, the integral state $\mX_3$ retains a weighted contribution from earlier delta states, including $\mS_1$, before the initial association was weakened. This retained history allows it to recover the first \textsf{No} answer.
This result also holds in other settings; we chose this simple case for clarity. The full proof is in Appendix~\ref{app:event-integral}.

\paragraph{Parametrization.}
We compute both gates introduced in \eqref{eq:integral_state} and \eqref{eq:readout-gate} from the normalized hidden state $\vh_t$ as
\begin{equation*}
  \vgamma_t = \exp\!\left[-\mathrm{softplus}\!\left(\mW_{\gamma}^{\uparrow}\mW_{\gamma}^{\downarrow}\vh_t+\vb_{\gamma}\right)\right], \quad
  \vlambda_t = \vone-\exp\!\left[-\mathrm{softplus}\!\left(\mW_{\lambda}^{\uparrow}\mW_{\lambda}^{\downarrow}\vh_t+\vb_{\lambda}\right)\right].
\end{equation*}
Following the low-rank parametrization of KDA, each projection uses a bottleneck rank small relative to the hidden dimension.
We initialize the gates so that $\vone-\vgamma_t$ and $\vlambda_t$ are both small.
This initialization places nearly all readout weight on the delta state $\mS_t$, while $\mX_t$ gradually accumulates a smoothed history of $\mS_t$.
The gates learn its retention time and readout weight.

\subsection{Hardware Efficient Training}\label{subsec:chunk}
Chunkwise form supports hardware-efficient training by parallelizing matrix operations within chunks while propagating recurrent states across chunks~\citep{yang2024gated,yang2024fla}. We derive this formulation for Integral DeltaNet.
For simplicity, assume $L=NC$ and index the $N$ chunks of size $C$ by $t=0,\ldots,N-1$.
A superscript $r\in\{1,\ldots,C\}$ denotes a position within chunk $[t]$.
We write $\square_{[t]}^r:=\square_{tC+r}$ for $\square\in\{\vq,\vk,\vv,\vo,\vlambda,\vgamma,\mS,\mX\}$.
For the recurrent states, we define $\square_{[t]}:=\square_{[t]}^0:=\square_{tC}$ for $\square\in\{\mS,\mX\}$, where $r=0$ marks the chunk start; hence $\square_{[t+1]}=\square_{[t]}^C$.

The delta state evolves according to a linear RNN. Within each chunk, the state at position $r$ is
\begin{equation*}
 \mS_{[t]}^r=\mS_{[t]}\mP_{[t]}^r+\mH_{[t]}^r,
 \qquad \mP_{[t]}^0=\mI,\quad \mH_{[t]}^0=\mzero,
 \qquad 0\le r\le C,
\end{equation*}
where $\mP_{[t]}^r\in\mathbb{R}^{d_k\times d_k}$ propagates the incoming state and
$\mH_{[t]}^r\in\mathbb{R}^{d_v\times d_k}$ collects the writes within the first $r$ positions.
Both operators depend only on the inputs and gates in the current chunk. A line of work~\citep{yang2024parallelizing,yang2025gated,team2025kimi,hatamizadeh2026gated} has developed efficient algorithms based on WY representations~\citep{bischof1987} and UT transforms~\citep{joffrain2006} to compute state updates and outputs without materializing all prefix matrices.

We express the integral state similarly using within-chunk retention weights for $0\le r\le C$:
\begin{equation*}
 \vomega_{[t]}^{0\to r}:=\prod_{s=1}^{r}\left(\vone-\vlambda_{[t]}^s\right),
 \qquad
 \vomega_{[t]}^{i\to r}:=\vlambda_{[t]}^i\odot
             \prod_{s=i+1}^{r}\left(\vone-\vlambda_{[t]}^s\right),
 \quad 1\le i\le r,
\end{equation*}
where products are elementwise and an empty product is $\vone$.
Define the retention-weighted operators:
\begin{equation*}
 \bar{\mP}_{[t]}^r:=\sum_{i=1}^{r}\mP_{[t]}^i\diag\!\left(\vomega_{[t]}^{i\to r}\right),
 \qquad
 \bar{\mH}_{[t]}^r:=\sum_{i=1}^{r}\mH_{[t]}^i\diag\!\left(\vomega_{[t]}^{i\to r}\right),
\end{equation*}
with empty sums equal to zero matrices of the corresponding dimensions.
Substituting the delta-state representation into the integral-state recurrence gives
\begin{equation*}
 \mX_{[t]}^r=\mS_{[t]}\bar{\mP}_{[t]}^r
       +\mX_{[t]}\diag\!\left(\vomega_{[t]}^{0\to r}\right)+\bar{\mH}_{[t]}^r.
\end{equation*}
Thus, the integral state combines retention-weighted prefix operators with the retained incoming integral state.
For the full-chunk operators, we omit the superscript $C$ and write $\square_{[t]}:=\square_{[t]}^C$ for $\square\in\{\mP,\mH,\bar{\mP},\bar{\mH}\}$.
The two states then propagate to the next chunk as
\begin{equation*}
  \mS_{[t+1]}=\mS_{[t]}\mP_{[t]}+\mH_{[t]},\quad
  \mX_{[t+1]}=\mS_{[t]}\bar{\mP}_{[t]}
             +\mX_{[t]}\diag\!\left(\vomega_{[t]}^{0\to C}\right)+\bar{\mH}_{[t]}.
\end{equation*}

The outputs similarly separate into contributions from the two incoming states and the writes within the chunk.
For $r=1,\ldots,C$, define the effective queries and the within-chunk output contribution as
\begin{equation*}
 \begin{gathered}
  \widetilde{\vq}_{\mathrm{S},[t]}^r
  :=\mP_{[t]}^r\left(\vgamma_{[t]}^r\odot\vq_{[t]}^r\right)
    +\bar{\mP}_{[t]}^r\left(\left(\vone-\vgamma_{[t]}^r\right)\odot\vq_{[t]}^r\right),\quad 
  \widetilde{\vq}_{\mathrm{X},[t]}^r
  :=\vomega_{[t]}^{0\to r}\odot\left(\vone-\vgamma_{[t]}^r\right)\odot\vq_{[t]}^r,\\
  \vz_{[t]}^r
  :=\mH_{[t]}^r\left(\vgamma_{[t]}^r\odot\vq_{[t]}^r\right)
    +\bar{\mH}_{[t]}^r\left(\left(\vone-\vgamma_{[t]}^r\right)\odot\vq_{[t]}^r\right).
 \end{gathered}
\end{equation*}
We stack $(\widetilde{\vq}_{a,[t]}^r)^\top$, $(\vz_{[t]}^r)^\top$, and $(\vo_{[t]}^r)^\top$ as the $r$-th rows of $\widetilde{\mQ}_{a,[t]}\in\mathbb{R}^{C\times d_k}$ and $\mZ_{[t]},\mO_{[t]}\in\mathbb{R}^{C\times d_v}$, respectively, for $a\in\{\mathrm{S},\mathrm{X}\}$.
The chunk output is then
\begin{equation*}
 \mO_{[t]}
 =\widetilde{\mQ}_{\mathrm{S},[t]}\mS_{[t]}^\top
  +\widetilde{\mQ}_{\mathrm{X},[t]}\mX_{[t]}^\top+\mZ_{[t]}.
\end{equation*}
We implement Integral DeltaNet kernels using this chunkwise form and FlashLinearAttention~\citep{yang2024fla}.
We compare training throughput and activation memory in Section~\ref{sec:exp}.

%% file: Sections/04Experiments.tex
\section{Experiments}\label{sec:exp}

We apply Integral DeltaNet to KDA and GDN-2, yielding \emph{Integral KDA} and \emph{Integral GDN-2}, respectively. 
We compare them with strong recurrent baselines Mamba-2, DeltaNet, GDN, KDA, and GDN-2, and with Transformer++~\citep{touvron2023llama}.
We pretrain all models from scratch with 340M/1.3B non-embedding parameters on 15B/100B FineWeb-Edu~\citep{penedo2024fineweb} tokens, using Llama-2 tokenization and 4K sequences.

Both scales use AdamW~\citep{loshchilov2019decoupled} ($\beta_1=0.9$, $\beta_2=0.95$, $\epsilon=10^{-8}$), 0.5M-token global batches, cosine learning-rate decay from $4\times10^{-4}$ to $4\times10^{-5}$, weight decay 0.1, and gradient clipping at 1.0.
We use 0.5B/1B tokens of linear warmup at 340M/1.3B.
We initialize rank-128 gate projections so that $\lambda_{t,j}\approx1-\gamma_{t,j}\approx1/1024$ for each key channel $j$. 

We evaluate these trained models on language modeling, commonsense reasoning, and real-world retrieval at both scales.
For tasks requiring longer contexts, we additionally evaluate the 1.3B models on Needle-in-a-Haystack~\citep{hsieh2024ruler}, LongBench~\citep{bai2024longbench}, and length extrapolation.
Finally, we evaluate training and decoding efficiency at the 1.3B scale. 
In the tables, we bold the better result in each baseline/Integral pair and mark the top two distinct scores among recurrent models with $\star$/$\dagger$ at each scale.

\paragraph{Language Modeling and Commonsense Reasoning.}
We use the LM Evaluation Harness~\citep{eval-harness} for evaluation.
We report perplexity (ppl) on WikiText-2~\citep{merity2017pointer} and LAMBADA~\citep{paperno2016lambada} and zero-shot accuracy (acc) on nine tasks (Table~\ref{tab:lm-commonsense}).
We use length-normalized accuracy ($\mathrm{acc}_n$) on selected tasks to reduce answer-length bias.
Both Integral variants show overall gains at both scales.

\begingroup
\begin{table}[ht]
  \centering
  \setlength{\tabcolsep}{4pt}
  \renewcommand{\arraystretch}{0.96}
  \setlength{\fboxsep}{1pt}
  \setlength{\aboverulesep}{2pt}
  \setlength{\belowrulesep}{3pt}
  \caption{Performance on language modeling and zero-shot commonsense reasoning.}
  \label{tab:lm-commonsense}
  \resizebox{\textwidth}{!}{%
  \begin{tabular}{@{}l@{\hspace{1em}}cc@{\hspace{1em}}ccccccccc@{\hspace{1em}}c@{}}
    \toprule[1pt]
    Model & Wiki. & LMB. & LMB. & PIQA & Hella. & Wino. & ARC-e & ARC-c & OBQA & SIQA & BoolQ & Avg. \\[2pt]
          & ppl $\downarrow$ & ppl $\downarrow$ & acc $\uparrow$ & acc $\uparrow$ & acc$_n$ $\uparrow$
          & acc $\uparrow$ & acc $\uparrow$ & acc$_n$ $\uparrow$ & acc$_n$ $\uparrow$ & acc $\uparrow$
          & acc $\uparrow$ & acc $\uparrow$ \\
    \midrule
    \multicolumn{13}{c}{\textbf{340M parameters / 15B tokens}} \\
    \midrule
Transformer++ & 26.37 & 36.20 & 33.1 & 65.8 & 39.3 & 51.1 & 59.0 & 27.6 & 33.6 & 38.1 & 59.5 & 45.2 \\
Mamba-2 & 27.95 & 36.89 & 30.1 & 66.4 & 39.5 & 50.8 & \second{58.2} & 27.0 & 33.8 & \second{38.9} & 55.3 & 44.4 \\
DeltaNet & 27.52 & 38.86 & 30.9 & 66.1 & 38.0 & 50.5 & 56.2 & 26.9 & 31.4 & 37.7 & 57.0 & 43.8 \\
GDN & 26.32 & 31.37 & 33.4 & 66.8 & 40.4 & 52.6 & 57.7 & 27.0 & \second{34.2} & 38.4 & 56.7 & 45.3 \\
\addlinespace[0.5pt]
\kdarow KDA & \best{\textbf{24.89}} & \second{26.16} & \second{36.4} & \second{66.9} & \best{\textbf{41.2}} & \best{\textbf{53.4}} & \best{\textbf{59.4}} & \second{27.3} & 33.6 & \best{\textbf{39.0}} & \second{57.9} & \second{46.1} \\
\kdarow Integral KDA & \second{24.97} & \best{\textbf{24.99}} & \best{\textbf{36.9}} & \best{\textbf{68.4}} & \second{40.7} & \second{52.9} & \second{58.2} & \best{\textbf{28.0}} & \best{\textbf{34.8}} & 38.8 & \best{\textbf{59.5}} & \best{\textbf{46.5}} \\
\addlinespace[0.5pt]
\gdntworow GDN-2 & \textbf{26.76} & 30.13 & 34.1 & \textbf{66.2} & 38.5 & 49.3 & 56.0 & \textbf{26.9} & 32.4 & 38.6 & \best{\textbf{59.5}} & 44.6 \\
\gdntworow Integral GDN-2 & 27.59 & \textbf{28.82} & \textbf{35.8} & 65.4 & \textbf{38.7} & \textbf{52.8} & \textbf{57.4} & 26.2 & \textbf{33.0} & \second{\textbf{38.9}} & 57.2 & \textbf{45.0} \\
    \midrule[1pt]
    \multicolumn{13}{c}{\textbf{1.3B parameters / 100B tokens}} \\
    \midrule
Transformer++ & 15.90 & 11.04 & 49.4 & 72.0 & 58.4 & 58.6 & 72.1 & 37.9 & 40.2 & 41.0 & 62.2 & 54.7 \\
Mamba-2 & 16.55 & 12.22 & 45.6 & 72.0 & 58.2 & 56.4 & \second{71.6} & 38.7 & \best{41.6} & 39.7 & \best{63.0} & 54.1 \\
DeltaNet & 15.73 & 11.51 & 47.3 & \second{73.2} & 58.1 & 56.7 & 71.0 & \second{40.0} & 41.0 & 40.4 & 60.7 & 54.3 \\
GDN & 16.08 & 10.98 & 48.5 & \second{73.2} & 58.7 & \second{59.4} & 71.3 & 39.3 & 40.8 & 41.0 & 59.9 & \second{54.7} \\
\addlinespace[0.5pt]
\kdarow KDA & \second{15.33} & \second{9.63} & 50.4 & \best{\textbf{73.9}} & \second{60.2} & 57.6 & 71.2 & 39.0 & \best{\textbf{41.6}} & 40.7 & 57.2 & 54.6 \\
\kdarow Integral KDA & \best{\textbf{15.00}} & \best{\textbf{9.16}} & \best{\textbf{52.7}} & \second{73.2} & \best{\textbf{60.7}} & \best{\textbf{61.1}} & \best{\textbf{72.3}} & \best{\textbf{40.6}} & 41.2 & \best{\textbf{42.5}} & \textbf{60.7} & \best{\textbf{56.1}} \\
\addlinespace[0.5pt]
\gdntworow GDN-2 & 16.34 & 10.56 & 49.3 & \textbf{71.9} & \textbf{57.9} & 57.6 & 70.0 & \textbf{39.3} & 39.8 & \second{\textbf{41.5}} & \second{\textbf{60.9}} & 54.2 \\
\gdntworow Integral GDN-2 & \textbf{15.92} & \textbf{10.31} & \second{\textbf{50.6}} & 71.7 & 57.8 & \textbf{59.2} & \textbf{71.5} & 38.1 & \second{\textbf{41.4}} & 41.1 & 59.6 & \textbf{54.6} \\
    \bottomrule[1pt]
  \end{tabular}%
  }
\end{table}
\endgroup

\begingroup
\paragraph{Real-World Retrieval.}
We evaluate document retrieval tasks~\citep{arora2024just} (Table~\ref{tab:jrt}).
SWDE~\citep{hao2011tree} and FDA~\citep{arora2023language} test extraction from webpages and medical reports respectively.
SQuAD~\citep{rajpurkar2018know} and TriviaQA~\citep{joshi2017triviaqa} test question answering alongside NQ~\citep{kwiatkowski2019natural} and DROP~\citep{dua2019drop}.
Both Integral variants improve average retrieval accuracy over their baselines at both scales.

\begin{table}[ht]
  \centering
  \small
  \setlength{\tabcolsep}{2.8pt}
  \renewcommand{\arraystretch}{1.12}
  \caption{Accuracy ($\%$, $\uparrow$) on six real-world retrieval tasks.}
  \label{tab:jrt}
  \resizebox{\textwidth}{!}{%
  \begin{tabular}{@{}l@{\hspace{1em}}*{6}{c}@{\hspace{0.8em}}c@{\hspace{1.5em}}*{6}{c}@{\hspace{0.8em}}c@{}}
    \toprule[1pt]
    & \multicolumn{7}{c}{\textbf{340M parameters / 15B tokens}} & \multicolumn{7}{c}{\textbf{1.3B parameters / 100B tokens}} \\
    \cmidrule(lr){2-8}\cmidrule(l){9-15}
    Model & SWDE & SQuAD & FDA & TriviaQA & NQ & DROP & \textbf{Avg.}
          & SWDE & SQuAD & FDA & TriviaQA & NQ & DROP & \textbf{Avg.} \\
    \midrule
Transformer++ & 31.8 & 34.2 & 48.2 & 48.4 & 17.3 & 19.4 & 33.2 & 48.2 & 45.3 & 65.7 & 62.8 & 26.7 & 25.1 & 45.6 \\
Mamba-2 & 6.6 & 24.0 & 3.7 & 42.9 & 10.5 & 14.9 & 17.1 & 20.7 & 36.5 & 13.6 & 59.4 & 20.1 & \second{21.1} & 28.6 \\
DeltaNet & 7.0 & 23.7 & 7.7 & 45.7 & 12.4 & 16.1 & 18.8 & 29.6 & 36.7 & 27.4 & 60.3 & 21.9 & 21.0 & 32.8 \\
GDN & 11.2 & 27.6 & 7.2 & 46.3 & 14.2 & 17.5 & 20.7 & \second{34.0} & \second{39.6} & 27.3 & 62.3 & \second{24.1} & 20.7 & 34.7 \\
\addlinespace[2pt]
\kdarow KDA & 10.2 & \second{28.6} & 8.1 & \second{47.8} & \second{14.4} & \second{\textbf{17.7}} & 21.1 & 32.7 & \best{\textbf{41.1}} & \second{30.2} & \best{\textbf{62.9}} & \second{24.1} & \best{\textbf{21.6}} & \second{35.4} \\
\kdarow Integral KDA & \best{\textbf{18.1}} & \best{\textbf{30.6}} & \best{\textbf{9.6}} & \best{\textbf{48.9}} & \best{\textbf{15.0}} & 17.3 & \best{\textbf{23.3}} & \best{\textbf{40.4}} & 39.5 & \best{\textbf{35.1}} & 62.4 & \best{\textbf{25.3}} & 20.9 & \best{\textbf{37.3}} \\
\addlinespace[2pt]
\gdntworow GDN-2 & 10.2 & 26.9 & 4.5 & \textbf{45.9} & 13.3 & \best{\textbf{18.3}} & 19.8 & 24.5 & 37.5 & 27.9 & 60.9 & 23.3 & 19.9 & 32.3 \\
\gdntworow Integral GDN-2 & \second{\textbf{15.7}} & \textbf{27.4} & \second{\textbf{8.9}} & 45.6 & \textbf{14.0} & 16.2 & \second{\textbf{21.3}} & \textbf{31.7} & \textbf{39.4} & \textbf{29.3} & \second{\textbf{62.6}} & \textbf{24.0} & \textbf{20.2} & \textbf{34.5} \\
    \bottomrule[1pt]
  \end{tabular}%
  }
\end{table}

\paragraph{Needle-in-a-Haystack (NIAH).}
We evaluate long-context retrieval using four synthetic Needle-in-a-Haystack tasks from RULER~\citep{hsieh2024ruler}.
We evaluate the 1.3B models on the same 500 examples per task and context length (Table~\ref{tab:niah}).
S-NIAH-1/2 retrieve a number from repetitive/natural text, respectively; S-NIAH-3 retrieves a UUID from natural text, and MK-NIAH-1 queries one of multiple key--value pairs.
Both Integral variants improve overall performance: Integral KDA matches or outperforms KDA in nearly all settings, with substantial gains on the challenging S-NIAH-3 at 4K and MK-NIAH-1 tasks.
Integral GDN-2 underperforms GDN-2 in some settings but improves S-NIAH-3 at 4K and achieves particularly large gains on MK-NIAH-1 at every tested length.
\begin{table}[ht]
  \centering
  \setlength{\tabcolsep}{4pt}
  \renewcommand{\arraystretch}{1.08}
  \caption{Accuracy ($\%$, $\uparrow$) on Needle-in-a-Haystack tasks.}
  \label{tab:niah}
  \resizebox{\textwidth}{!}{%
  \begin{tabular}{@{}l@{\hspace{1em}}cccc@{\hspace{1em}}cccc@{\hspace{1em}}ccc@{\hspace{1em}}ccc@{}}
    \toprule[1pt]
    & \multicolumn{4}{c}{S-NIAH-1} & \multicolumn{4}{c}{S-NIAH-2}
    & \multicolumn{3}{c}{S-NIAH-3} & \multicolumn{3}{c}{MK-NIAH-1} \\
    \cmidrule(lr){2-5}\cmidrule(lr){6-9}\cmidrule(lr){10-12}\cmidrule(l){13-15}
    Model & 1K & 2K & 4K & 8K & 1K & 2K & 4K & 8K
          & 1K & 2K & 4K & 1K & 2K & 4K \\
    \midrule
Transformer++ & 100.0 & 100.0 & 98.4 & 0.0 & 100.0 & 100.0 & 86.2 & 0.0 & 100.0 & 99.4 & 69.8 & 98.4 & 94.4 & 77.0 \\
Mamba-2 & \best{100.0} & \second{99.8} & 69.0 & 21.4 & \best{100.0} & 98.6 & 65.8 & 23.8 & 63.4 & 46.6 & 13.2 & 29.2 & 21.2 & 18.4 \\
DeltaNet & \best{100.0} & \best{100.0} & \best{100.0} & \best{100.0} & \best{100.0} & 97.4 & 80.6 & \second{47.6} & \best{95.2} & 65.2 & 21.0 & 52.8 & 33.4 & 30.4 \\
GDN & \best{100.0} & 99.2 & 91.6 & 63.6 & 88.8 & \best{99.6} & \second{89.8} & 28.0 & 90.8 & 63.2 & 20.0 & 48.8 & 39.0 & 25.0 \\
\addlinespace[2pt]
\kdarow KDA & \best{\textbf{100.0}} & \best{\textbf{100.0}} & 99.6 & 78.4 & \best{\textbf{100.0}} & 79.8 & 70.0 & 32.8 & 90.4 & \textbf{68.8} & 24.8 & 47.2 & 37.0 & 20.8 \\
\kdarow Integral KDA & \best{\textbf{100.0}} & \best{\textbf{100.0}} & \best{\textbf{100.0}} & \textbf{92.4} & \best{\textbf{100.0}} & \second{\textbf{99.2}} & \best{\textbf{99.6}} & \best{\textbf{49.2}} & \textbf{91.8} & 58.6 & \best{\textbf{41.6}} & \second{\textbf{54.0}} & \second{\textbf{50.4}} & \second{\textbf{35.6}} \\
\addlinespace[2pt]
\gdntworow GDN-2 & \best{\textbf{100.0}} & \best{\textbf{100.0}} & \second{\textbf{99.8}} & \second{\textbf{97.8}} & \second{99.6} & 93.8 & \textbf{74.2} & \textbf{30.8} & \second{\textbf{94.0}} & \best{\textbf{80.0}} & 30.0 & 45.2 & 32.8 & 23.8 \\
\gdntworow Integral GDN-2 & \best{\textbf{100.0}} & \second{99.8} & 98.8 & 85.8 & \best{\textbf{100.0}} & \textbf{94.6} & 65.2 & 29.4 & 93.4 & \second{77.4} & \second{\textbf{34.6}} & \best{\textbf{76.4}} & \best{\textbf{67.2}} & \best{\textbf{39.8}} \\
    \bottomrule[1pt]
  \end{tabular}%
  }
\end{table}

\paragraph{LongBench.}
We assess long-context understanding using the 1.3B models on English LongBench tasks~\citep{bai2024longbench} covering question answering, summarization, few-shot learning, and code completion.
We use the LM Evaluation Harness~\citep{eval-harness} with greedy decoding, a $16$K-token context budget, and default prompts and truncation, reporting the mean of official task scores.
Both Integral variants achieve strong overall performance on LongBench (Table~\ref{tab:longbench}).

\begin{table}[ht]
  \centering
  \small
  \setlength{\tabcolsep}{2.6pt}
  \renewcommand{\arraystretch}{1.12}
  \caption{Scores ($\uparrow$) on LongBench tasks.}
  \label{tab:longbench}
  \resizebox{\textwidth}{!}{%
  \begin{tabular}{@{}l@{\hspace{1em}}*{14}{c}@{\hspace{0.8em}}c@{}}
    \toprule[1pt]
    & \multicolumn{3}{c}{Single-Doc QA} & \multicolumn{3}{c}{Multi-Doc QA} & \multicolumn{3}{c}{Summarization} & \multicolumn{3}{c}{Few-shot} & \multicolumn{2}{c}{Code} & \\
    \cmidrule(lr){2-4}\cmidrule(lr){5-7}\cmidrule(lr){8-10}\cmidrule(lr){11-13}\cmidrule(lr){14-15}
    Model & NQA & QQA & MFQ & HQA & 2WM & Mus & GvR & QMS & MNs & TRC & TQA & SSM & LCC & RBP & \textbf{Avg.} \\
    \midrule
Transformer++ & 0.5 & 2.3 & 5.3 & 1.1 & 2.2 & 0.4 & 5.3 & 5.8 & 7.5 & 11.5 & 6.6 & 3.9 & 17.4 & 8.4 & 5.6 \\
Mamba-2 & 2.0 & 4.3 & 12.5 & 4.1 & 8.1 & \second{3.1} & 6.6 & 16.6 & 8.7 & 31.0 & 22.8 & 20.5 & 11.2 & 17.8 & 12.1 \\
DeltaNet & 2.2 & 3.9 & 13.6 & 5.3 & 8.1 & 2.4 & 4.9 & \second{17.5} & 10.4 & 37.0 & 19.8 & \second{27.6} & 15.7 & 18.3 & 13.3 \\
GDN & 2.1 & \second{5.1} & \second{13.7} & 5.1 & 8.4 & \best{3.3} & 6.8 & \best{18.3} & 12.3 & 26.0 & 20.7 & 23.3 & \second{17.7} & 19.4 & 13.0 \\
\addlinespace[2pt]
\kdarow KDA & 2.0 & 4.3 & \textbf{13.0} & \textbf{5.1} & \textbf{8.4} & \second{\textbf{3.1}} & \second{\textbf{10.0}} & 16.2 & 9.6 & 29.5 & 20.5 & 27.5 & \best{\textbf{21.9}} & \best{\textbf{24.0}} & 13.9 \\
\kdarow Integral KDA & \best{\textbf{2.6}} & \textbf{4.4} & 9.7 & \textbf{5.1} & 8.1 & 2.9 & 6.6 & \textbf{16.4} & \textbf{11.8} & \best{\textbf{52.5}} & \best{\textbf{27.3}} & \best{\textbf{28.1}} & 16.3 & 18.0 & \best{\textbf{15.0}} \\
\addlinespace[2pt]
\gdntworow GDN-2 & 2.2 & \best{\textbf{5.2}} & 13.1 & \second{5.5} & \best{\textbf{9.2}} & \best{\textbf{3.3}} & 9.8 & \textbf{16.7} & \second{12.4} & \second{\textbf{42.0}} & \second{\textbf{25.4}} & 24.0 & 15.4 & \second{\textbf{19.7}} & 14.6 \\
\gdntworow Integral GDN-2 & \second{\textbf{2.5}} & \best{\textbf{5.2}} & \best{\textbf{13.9}} & \best{\textbf{5.9}} & \second{9.1} & 2.8 & \best{\textbf{12.5}} & 16.1 & \best{\textbf{13.0}} & 40.5 & 24.8 & \textbf{27.4} & \textbf{15.9} & 17.1 & \second{\textbf{14.8}} \\
    \bottomrule[1pt]
  \end{tabular}%
  }
\end{table}

\endgroup

\paragraph{Length Extrapolation.}
We evaluate how well Integral DeltaNet generalizes beyond the 4K training length compared with other recurrent models.
We measure perplexity of the 1.3B models up to 20K on six datasets: GovReport~\citep{huang2021efficient}, QMSum~\citep{zhong2021qmsum}, NarrativeQA~\citep{kocisky2018narrativeqa}, Qasper~\citep{dasigi2021qasper}, CodeParrot~\citep{vonwerra2021codeparrot}, and PG19~\citep{rae2019compressive}.
Both Integral variants outperform their respective baselines across all six datasets at every evaluated length beyond the 4K training context.
Together, they achieve the lowest perplexity among all compared models at every evaluated length on every dataset: Integral KDA leads on GovReport, QMSum, Qasper, and CodeParrot, while Integral GDN-2 leads on NarrativeQA and PG19 (Figure~\ref{fig:extrapolation}).

\input{Sections/fig_extrapolation}
\paragraph{Training Throughput and Memory Cost.}
We compare recurrent baselines using Flash Linear Attention (FLA) kernels, Transformer++ using FlashAttention-2 and FlashAttention-4, and Integral using our FLA-based implementation.
We provide two implementations: \emph{throughput-first} stores intermediates for reuse, while \emph{memory-first} recomputes them during backpropagation.
We measure training throughput and activation memory of 1.3B models on a B300 at a fixed token budget without model-level activation checkpointing (Figure~\ref{fig:training-efficiency}).
Throughput-first retains throughput close to the backbones, while memory-first substantially reduces activation memory with a moderate slowdown.\footnote{The FLA kernels used here recompute kernel-internal intermediates; our memory-first implementation recomputes the entire mixer, allowing Integral to use less activation memory than the backbones.}\par

\begingroup
\begin{figure}[ht]

  \centering
  \resizebox{\linewidth}{!}{%
  \begin{tabular}{@{}c@{\hspace{0.01\textwidth}}c@{}}
    \textbf{(a) Training throughput} & \textbf{(b) Activation memory} \\[2pt]
    \input{Sections/fig_throughput_gdn} & \input{Sections/fig_activation_memory}
  \end{tabular}%
  }
  \par
  \pgfplotslegendfromname{training-efficiency-legend}
  \caption{Training throughput and activation memory. T: throughput-first; M: memory-first.}
  \label{fig:training-efficiency}
  \label{fig:throughput}
  \label{fig:actmem}
\par

\end{figure}
\endgroup

\paragraph{Decoding Memory and Latency.}
We compare decoding memory and per-token latency of 1.3B models on a B300 with batch size $8$ across context lengths (Table~\ref{tab:decode-cost}).
Memory is computed from the architecture; latency is measured with CUDA graph replay, excluding prefill.
Both costs remain nearly constant for recurrent models, while Transformer++ requires increasing memory and computation as its KV cache grows.
The integral state adds a small memory overhead and increases per-token latency while preserving this context-length scaling.\par

\begin{table}[ht]
  \centering
  \small
  \setlength{\tabcolsep}{5pt}
  \renewcommand{\arraystretch}{1.0}
  \caption{Decoding memory and per-token latency across context lengths.}
  \label{tab:decode-cost}
  \begin{tabular}{@{}l@{\hspace{1.1em}} rrr @{\hspace{1.6em}} rrr@{}}
    \toprule
    & \multicolumn{3}{c}{Memory (GiB)} & \multicolumn{3}{c}{Latency (ms)} \\
    \cmidrule(lr){2-4}\cmidrule(l){5-7}
    Model & $4$K & $16$K & $64$K & $4$K & $16$K & $64$K \\
    \midrule
    Mamba-2 & 2.85 & 2.85 & 2.85 & 2.15 & 2.15 & 2.15 \\
    DeltaNet & 2.87 & 2.87 & 2.87 & 2.61 & 2.60 & 2.60 \\
    GDN & 2.94 & 2.94 & 2.94 & 2.50 & 2.51 & 2.50 \\
    \addlinespace[2pt]
    \kdarow KDA & 2.85 & 2.85 & 2.85 & 3.34 & 3.34 & 3.34 \\
    \kdarow Integral KDA & 2.99 & 2.99 & 2.99 & 4.63 & 4.63 & 4.63 \\
    \addlinespace[2pt]
    \gdntworow GDN-2 & 2.85 & 2.85 & 2.85 & 3.19 & 3.19 & 3.19 \\
    \gdntworow Integral GDN-2 & 2.99 & 2.99 & 2.99 & 4.79 & 4.79 & 4.79 \\
    \midrule
    Transformer++ & 7.77 & 22.95 & 83.70 & 5.18 & 12.74 & 42.59 \\
    \bottomrule
  \end{tabular}
\end{table}

%% file: Sections/fig_extrapolation.tex
\definecolor{vizA}{HTML}{2A78D6}
\definecolor{vizB}{HTML}{EB6834}
\definecolor{vizC}{HTML}{1BAF7A}
\definecolor{vizD}{HTML}{EDA100}
\definecolor{vizE}{HTML}{E87BA4}
\definecolor{vizF}{HTML}{008300}

\begingroup
\begin{figure}[ht]
\pgfplotsset{
  extrap recurrent/.style={solid, mark=*, line width=0.65pt,
    mark size=1.0pt, mark options={solid}},
  extrap integral recurrent/.style={dashed, mark=*, line width=0.8pt,
    mark size=1.0pt, mark options={solid, fill=white, line width=0.65pt}},
}
  \centering
  \begin{tikzpicture}
    \begin{groupplot}[
      paperplot,
      group style={group size=6 by 1, horizontal sep=0.85cm,
                   vertical sep=1.0cm,
                   ylabels at=edge left, xlabels at=edge bottom},
      scale only axis, width=1.9cm, height=1.9cm,
      xmode=log,
      xmin=3809, xmax=21913,
      xtick={4096,8192,20480}, xticklabels={$4\mathrm{k}$,$8\mathrm{k}$,$20\mathrm{k}$},
      minor xtick={},
      xmajorgrids=false, ymajorgrids=true,
      grid style={gray!15, line width=0.25pt},
      title style={yshift=-3pt},
      legend style={draw=none, fill=white,
                    legend columns=-1,
                    /tikz/every even column/.append style={column sep=7pt}},
      legend cell align=left,
      every axis plot/.append style={mark repeat=4},
      ylabel={Perplexity},
      xlabel={Length},
    ]
      \nextgroupplot[
        title={GovReport},
        legend to name={extrapLegend},
        ymin=5.6, ymax=6.6, ytick={5.6,5.9,6.2,6.5},
      ]
        \addplot[forget plot, color=vizB, mark=square*, line width=0.55pt, mark size=0.9pt] coordinates {(4096,6.277) (6144,6.111) (8192,6.079) (10240,6.171) (12288,6.272) (14336,6.329) (16384,6.376) (18432,6.410) (20480,6.426)};

        \addplot[forget plot, color=vizC, mark=diamond*, line width=0.55pt, mark size=0.9pt] coordinates {(4096,6.168) (6144,5.976) (8192,5.924) (10240,5.991) (12288,6.052) (14336,6.082) (16384,6.100) (18432,6.122) (20480,6.125)};

        \addplot[forget plot, color=vizD, mark=triangle*, line width=0.55pt, mark size=1.0pt] coordinates {(4096,6.104) (6144,5.923) (8192,5.868) (10240,5.935) (12288,5.994) (14336,6.019) (16384,6.031) (18432,6.046) (20480,6.039)};

        \addplot[forget plot, color=vizF, extrap recurrent] coordinates {(4096,6.216) (6144,6.031) (8192,5.987) (10240,6.053) (12288,6.106) (14336,6.128) (16384,6.133) (18432,6.138) (20480,6.123)};

        \addplot[forget plot, color=vizE, extrap recurrent] coordinates {(4096,5.989) (6144,5.799) (8192,5.748) (10240,5.808) (12288,5.858) (14336,5.877) (16384,5.881) (18432,5.887) (20480,5.872)};

        \addplot[forget plot, color=vizE!80!black, extrap integral recurrent] coordinates {(4096,5.931) (6144,5.749) (8192,5.693) (10240,5.751) (12288,5.796) (14336,5.810) (16384,5.814) (18432,5.820) (20480,5.802)};

        \addplot[forget plot, color=vizF!80!black, extrap integral recurrent] coordinates {(4096,6.176) (6144,5.986) (8192,5.922) (10240,5.982) (12288,6.035) (14336,6.050) (16384,6.055) (18432,6.060) (20480,6.047)};

        \addlegendimage{color=vizB, mark=square*, line width=0.55pt, mark size=0.9pt}
        \addlegendentry{Mamba-2}
        \addlegendimage{color=vizC, mark=diamond*, line width=0.55pt, mark size=0.9pt}
        \addlegendentry{DeltaNet}
        \addlegendimage{color=vizD, mark=triangle*, line width=0.55pt, mark size=1.0pt}
        \addlegendentry{GDN}
        \addlegendimage{empty legend}
        \addlegendentry{}
        \addlegendimage{color=vizE, extrap recurrent}
        \addlegendentry{KDA}
        \addlegendimage{color=vizE!80!black, extrap integral recurrent}
        \addlegendentry{Integral KDA}
        \addlegendimage{color=vizF, extrap recurrent}
        \addlegendentry{GDN-2}
        \addlegendimage{color=vizF!80!black, extrap integral recurrent}
        \addlegendentry{Integral GDN-2}
      \nextgroupplot[
        title={QMSum},
        ymin=12, ymax=15.5, ytick={12,13,14,15},
      ]
        \addplot[forget plot, color=vizB, mark=square*, line width=0.55pt, mark size=0.9pt] coordinates {(4096,15.135) (6144,14.592) (8192,14.263) (10240,14.142) (12288,14.250) (14336,14.346) (16384,14.456) (18432,14.574) (20480,14.677)};
        \addplot[forget plot, color=vizC, mark=diamond*, line width=0.55pt, mark size=0.9pt] coordinates {(4096,14.675) (6144,14.055) (8192,13.660) (10240,13.449) (12288,13.452) (14336,13.468) (16384,13.475) (18432,13.498) (20480,13.527)};
        \addplot[forget plot, color=vizD, mark=triangle*, line width=0.55pt, mark size=1.0pt] coordinates {(4096,14.791) (6144,14.191) (8192,13.783) (10240,13.541) (12288,13.512) (14336,13.481) (16384,13.446) (18432,13.426) (20480,13.410)};
        \addplot[forget plot, color=vizF, extrap recurrent] coordinates {(4096,14.629) (6144,14.030) (8192,13.605) (10240,13.361) (12288,13.333) (14336,13.304) (16384,13.255) (18432,13.238) (20480,13.223)};
        \addplot[forget plot, color=vizE, extrap recurrent] coordinates {(4096,14.466) (6144,13.855) (8192,13.443) (10240,13.201) (12288,13.154) (14336,13.115) (16384,13.069) (18432,13.047) (20480,13.019)};
        \addplot[forget plot, color=vizE!80!black, extrap integral recurrent] coordinates {(4096,13.943) (6144,13.322) (8192,12.899) (10240,12.644) (12288,12.597) (14336,12.557) (16384,12.509) (18432,12.477) (20480,12.450)};
        \addplot[forget plot, color=vizF!80!black, extrap integral recurrent] coordinates {(4096,14.347) (6144,13.747) (8192,13.321) (10240,13.072) (12288,13.021) (14336,12.984) (16384,12.937) (18432,12.898) (20480,12.872)};
      \nextgroupplot[
        title={NarrativeQA},
        ymin=15, ymax=30, ytick={15,20,25,30},
      ]
        \addplot[forget plot, color=vizB, mark=square*, line width=0.55pt, mark size=0.9pt] coordinates {(4096,22.500) (6144,22.364) (8192,22.317) (10240,22.455) (12288,22.547) (14336,22.508) (16384,22.464) (18432,22.518) (20480,22.486)};
        \addplot[forget plot, color=vizC, mark=diamond*, line width=0.55pt, mark size=0.9pt] coordinates {(4096,17.393) (6144,17.006) (8192,16.921) (10240,16.995) (12288,17.053) (14336,17.030) (16384,17.032) (18432,17.127) (20480,17.177)};
        \addplot[forget plot, color=vizD, mark=triangle*, line width=0.55pt, mark size=1.0pt] coordinates {(4096,29.639) (6144,29.545) (8192,29.323) (10240,29.553) (12288,29.571) (14336,29.354) (16384,29.032) (18432,29.011) (20480,28.893)};
        \addplot[forget plot, color=vizF, extrap recurrent] coordinates {(4096,20.258) (6144,19.875) (8192,19.747) (10240,19.749) (12288,19.725) (14336,19.640) (16384,19.575) (18432,19.591) (20480,19.541)};
        \addplot[forget plot, color=vizE, extrap recurrent] coordinates {(4096,18.358) (6144,17.991) (8192,17.813) (10240,17.819) (12288,17.794) (14336,17.687) (16384,17.621) (18432,17.610) (20480,17.560)};
        \addplot[forget plot, color=vizE!80!black, extrap integral recurrent] coordinates {(4096,18.176) (6144,17.756) (8192,17.559) (10240,17.554) (12288,17.509) (14336,17.389) (16384,17.297) (18432,17.277) (20480,17.208)};
        \addplot[forget plot, color=vizF!80!black, extrap integral recurrent] coordinates {(4096,16.944) (6144,16.444) (8192,16.298) (10240,16.310) (12288,16.301) (14336,16.210) (16384,16.161) (18432,16.180) (20480,16.147)};
      \nextgroupplot[
        title={Qasper},
        ymin=10, ymax=12.5, ytick={10,10.8,11.6,12.4},
      ]
        \addplot[forget plot, color=vizB, mark=square*, line width=0.55pt, mark size=0.9pt] coordinates {(4096,11.641) (6144,11.627) (8192,11.744) (10240,11.805) (12288,11.737) (14336,11.855) (16384,11.916) (18432,11.949) (20480,12.038)};
        \addplot[forget plot, color=vizC, mark=diamond*, line width=0.55pt, mark size=0.9pt] coordinates {(4096,10.991) (6144,10.914) (8192,10.974) (10240,10.982) (12288,10.879) (14336,10.920) (16384,10.926) (18432,10.928) (20480,10.972)};
        \addplot[forget plot, color=vizD, mark=triangle*, line width=0.55pt, mark size=1.0pt] coordinates {(4096,11.024) (6144,10.958) (8192,11.026) (10240,11.033) (12288,10.916) (14336,10.954) (16384,10.944) (18432,10.910) (20480,10.922)};
        \addplot[forget plot, color=vizF, extrap recurrent] coordinates {(4096,11.224) (6144,11.128) (8192,11.186) (10240,11.188) (12288,11.061) (14336,11.052) (16384,11.011) (18432,10.970) (20480,10.976)};
        \addplot[forget plot, color=vizE, extrap recurrent] coordinates {(4096,10.659) (6144,10.555) (8192,10.575) (10240,10.563) (12288,10.435) (14336,10.429) (16384,10.391) (18432,10.355) (20480,10.353)};
        \addplot[forget plot, color=vizE!80!black, extrap integral recurrent] coordinates {(4096,10.554) (6144,10.454) (8192,10.482) (10240,10.465) (12288,10.330) (14336,10.312) (16384,10.267) (18432,10.227) (20480,10.217)};
        \addplot[forget plot, color=vizF!80!black, extrap integral recurrent] coordinates {(4096,11.057) (6144,10.954) (8192,10.980) (10240,10.964) (12288,10.827) (14336,10.814) (16384,10.760) (18432,10.714) (20480,10.708)};
      \nextgroupplot[
        title={CodeParrot},
        ymin=8, ymax=14, ytick={8,10,12,14},
      ]
        \addplot[forget plot, color=vizB, mark=square*, line width=0.55pt, mark size=0.9pt] coordinates {(4096,12.867) (6144,12.388) (8192,12.624) (10240,12.825) (12288,13.264) (14336,13.231) (16384,13.191) (18432,13.192) (20480,13.256)};
        \addplot[forget plot, color=vizC, mark=diamond*, line width=0.55pt, mark size=0.9pt] coordinates {(4096,10.953) (6144,10.354) (8192,10.470) (10240,10.598) (12288,10.970) (14336,10.978) (16384,11.012) (18432,11.089) (20480,11.235)};
        \addplot[forget plot, color=vizD, mark=triangle*, line width=0.55pt, mark size=1.0pt] coordinates {(4096,11.785) (6144,11.044) (8192,11.191) (10240,11.371) (12288,11.673) (14336,11.558) (16384,11.478) (18432,11.480) (20480,11.509)};
        \addplot[forget plot, color=vizF, extrap recurrent] coordinates {(4096,9.859) (6144,9.331) (8192,9.440) (10240,9.594) (12288,9.889) (14336,9.818) (16384,9.784) (18432,9.805) (20480,9.820)};
        \addplot[forget plot, color=vizE, extrap recurrent] coordinates {(4096,10.772) (6144,10.215) (8192,10.270) (10240,10.359) (12288,10.652) (14336,10.582) (16384,10.532) (18432,10.529) (20480,10.574)};
        \addplot[forget plot, color=vizE!80!black, extrap integral recurrent] coordinates {(4096,9.447) (6144,8.834) (8192,8.855) (10240,8.945) (12288,9.201) (14336,9.139) (16384,9.092) (18432,9.087) (20480,9.116)};
        \addplot[forget plot, color=vizF!80!black, extrap integral recurrent] coordinates {(4096,9.907) (6144,9.299) (8192,9.301) (10240,9.393) (12288,9.670) (14336,9.629) (16384,9.609) (18432,9.624) (20480,9.668)};
      \nextgroupplot[
        title={PG19},
        ymin=14, ymax=19.2, ytick={14,15.5,17,18.5},
      ]
        \addplot[forget plot, color=vizB, mark=square*, line width=0.55pt, mark size=0.9pt] coordinates {(4096,18.902) (6144,18.785) (8192,18.621) (10240,18.372) (12288,18.298) (14336,18.320) (16384,18.366) (18432,18.443) (20480,18.432)};
        \addplot[forget plot, color=vizC, mark=diamond*, line width=0.55pt, mark size=0.9pt] coordinates {(4096,16.312) (6144,16.073) (8192,15.871) (10240,15.697) (12288,15.639) (14336,15.689) (16384,15.747) (18432,15.821) (20480,15.847)};
        \addplot[forget plot, color=vizD, mark=triangle*, line width=0.55pt, mark size=1.0pt] coordinates {(4096,17.984) (6144,17.774) (8192,17.620) (10240,17.389) (12288,17.297) (14336,17.309) (16384,17.355) (18432,17.400) (20480,17.368)};
        \addplot[forget plot, color=vizF, extrap recurrent] coordinates {(4096,18.033) (6144,17.843) (8192,17.683) (10240,17.480) (12288,17.403) (14336,17.430) (16384,17.474) (18432,17.517) (20480,17.491)};
        \addplot[forget plot, color=vizE, extrap recurrent] coordinates {(4096,16.112) (6144,15.924) (8192,15.762) (10240,15.564) (12288,15.469) (14336,15.467) (16384,15.502) (18432,15.539) (20480,15.499)};
        \addplot[forget plot, color=vizE!80!black, extrap integral recurrent] coordinates {(4096,15.851) (6144,15.598) (8192,15.387) (10240,15.150) (12288,15.045) (14336,15.028) (16384,15.041) (18432,15.061) (20480,15.005)};
        \addplot[forget plot, color=vizF!80!black, extrap integral recurrent] coordinates {(4096,15.624) (6144,15.314) (8192,15.107) (10240,14.898) (12288,14.802) (14336,14.791) (16384,14.794) (18432,14.830) (20480,14.779)};
    \end{groupplot}
  \end{tikzpicture}

  \pgfplotslegendfromname{extrapLegend}
  \caption{Length extrapolation of 1.3B models trained at 4K (perplexity, $\downarrow$).}
  \label{fig:extrapolation}
\end{figure}
\endgroup

%% file: Sections/fig_throughput_gdn.tex
\definecolor{vizA}{HTML}{2A78D6}
\definecolor{vizB}{HTML}{EB6834}
\definecolor{vizC}{HTML}{1BAF7A}
\definecolor{vizD}{HTML}{EDA100}
\definecolor{vizE}{HTML}{E87BA4}
\definecolor{vizF}{HTML}{008300}

  \begin{tikzpicture}
    \begin{axis}[
      paperplot,
      width=0.49\textwidth, height=6.0cm,
      xlabel={Sequence length $\times$ batch size},
      xticklabel style={font=\rmfamily\fontsize{7}{8.4}\selectfont},
      ylabel={Throughput (Kt/s)},
      xtick={1,2,3,4,5},
      xticklabels={$2\mathrm{K}\times16$,$4\mathrm{K}\times8$,$8\mathrm{K}\times4$,$16\mathrm{K}\times2$,$32\mathrm{K}\times1$},
      xmin=0.85, xmax=5.15, ymin=0, ymax=80,
      ytick={0,20,40,60,80},
      grid=major,
      grid style={gray!20, line width=0.3pt},
      legend to name=training-efficiency-legend,
      legend style={draw=none, fill=none,
                    font=\rmfamily\fontsize{7}{8.4}\selectfont,
                    legend columns=3,
                    inner sep=1pt, row sep=0pt,
                    nodes={inner sep=0.2pt, text height=5.3pt, text depth=1.2pt},
                    /tikz/every even column/.append style={column sep=5pt}},
      legend cell align=left,
      legend image post style={xscale=1.3},
    ]
      \addplot[color=vizA, mark=*]            coordinates {(1,76.1) (2,73.4) (3,68.7) (4,60.9) (5,50.3)};
      \addlegendentry{Transformer++ (FA4)}
      \addplot[color=vizA!55, mark=o, mark options={solid}]
                                              coordinates {(1,70.8) (2,63.8) (3,52.9) (4,39.5) (5,26.1)};
      \addlegendentry{Transformer++ (FA2)}
      \addplot[color=vizB, mark=square*]      coordinates {(1,71.7) (2,71.6) (3,71.7) (4,71.4) (5,71.5)};
      \addlegendentry{Mamba-2}
      \addplot[color=vizC, mark=triangle*]    coordinates {(1,68.1) (2,68.1) (3,67.8) (4,66.8) (5,66.0)};
      \addlegendentry{DeltaNet}
      \addplot[color=vizD, mark=diamond*]     coordinates {(1,58.5) (2,57.9) (3,56.9) (4,57.0) (5,53.5)};
      \addlegendentry{GDN}
      \addlegendimage{empty legend}
      \addlegendentry{}
      \addplot[color=vizE, solid, line width=0.9pt, mark=pentagon*, mark size=2.0pt]    coordinates {(1,58.1) (2,58.0) (3,57.6) (4,56.5) (5,54.5)};
      \addlegendentry{KDA}
      \addplot[color=vizE!80!black, line width=1.2pt, dash pattern=on 4pt off 2.2pt,
               mark=triangle*, mark size=2.4pt, mark options={solid, fill=vizE!80!black}]
                                              coordinates {(1,51.2) (2,52.8) (3,52.8) (4,51.2) (5,47.3)};
      \addlegendentry{Integral KDA (T)}
      \addplot[color=vizE!80!black, densely dotted, line width=1.0pt,
               mark=triangle, mark size=2.4pt, mark options={solid, fill=white}]
                                              coordinates {(1,45.0) (2,46.5) (3,46.5) (4,44.2) (5,40.0)};
      \addlegendentry{Integral KDA (M)}
      \addplot[color=vizF, solid, line width=0.9pt, mark=otimes*, mark size=2.0pt]      coordinates {(1,56.9) (2,56.8) (3,56.4) (4,54.6) (5,53.9)};
      \addlegendentry{GDN-2}
      \addplot[color=vizF!80!black, line width=1.2pt, dash pattern=on 4pt off 2.2pt,
               mark=square*, mark size=2.1pt, mark options={solid, fill=vizF!80!black}]
                                              coordinates {(1,51.8) (2,52.8) (3,52.8) (4,51.2) (5,47.5)};
      \addlegendentry{Integral GDN-2 (T)}
      \addplot[color=vizF!80!black, densely dotted, line width=1.0pt,
               mark=square, mark size=2.1pt, mark options={solid, fill=white}]
                                              coordinates {(1,44.9) (2,46.0) (3,45.3) (4,43.7) (5,39.5)};
      \addlegendentry{Integral GDN-2 (M)}
    \end{axis}
  \end{tikzpicture}

%% file: Sections/fig_activation_memory.tex
\definecolor{vizA}{HTML}{2A78D6}
\definecolor{vizB}{HTML}{EB6834}
\definecolor{vizC}{HTML}{1BAF7A}
\definecolor{vizD}{HTML}{EDA100}
\definecolor{vizE}{HTML}{E87BA4}
\definecolor{vizF}{HTML}{008300}

  \begin{tikzpicture}
    \begin{axis}[
      paperplot,
      scale only axis,
      width=0.225\textwidth, height=4.5cm,
      xbar, bar width=8pt, bar shift=0pt,
      xlabel={Activation memory (GiB)},
      xmin=0, xmax=215,
      xtick={0,50,100,150},
      symbolic y coords={tpp,m2,dn,gdn,kda,ikdaF,ikdaM,gdn2,igdn2F,igdn2M},
      ytick={tpp,m2,dn,gdn,kda,ikdaF,ikdaM,gdn2,igdn2F,igdn2M},
      yticklabels={{Transformer++ (FA4, FA2)},Mamba-2,DeltaNet,GDN,KDA,%
                   {Integral KDA (T)},{Integral KDA (M)},GDN-2,%
                   {Integral GDN-2 (T)},{Integral GDN-2 (M)}},
      yticklabel style={font=\rmfamily\fontsize{7}{8.4}\selectfont, align=right},
      xticklabel style={font=\rmfamily\fontsize{7}{8.4}\selectfont},
      axis on top=false,
      y dir=reverse,
      enlarge y limits=0.06,
      xmajorgrids=true, ymajorgrids=false,
      grid style={gray!20, line width=0.3pt},
      nodes near coords,
      nodes near coords style={
        font=\rmfamily\fontsize{7}{8.4}\selectfont,
        anchor=west, xshift=1pt, text=black,
        /pgf/number format/.cd, fixed, fixed zerofill, precision=1},
      every axis plot/.append style={line width=0.6pt},
    ]
      \addplot[fill=vizA,    draw=vizA] coordinates {( 58.8,tpp)};
      \addplot[fill=vizB,    draw=vizB] coordinates {( 54.8,m2)};
      \addplot[fill=vizC,    draw=vizC] coordinates {( 72.7,dn)};
      \addplot[fill=vizD,    draw=vizD] coordinates {( 73.9,gdn)};
      \addplot[fill=vizE,    draw=vizE] coordinates {( 81.7,kda)};
      \addplot[fill=vizE!18, draw=vizE] coordinates {(159.3,ikdaF)};
      \addplot[fill=white,   draw=vizE, line width=1.1pt] coordinates {( 49.8,ikdaM)};
      \addplot[fill=vizF,    draw=vizF] coordinates {(101.5,gdn2)};
      \addplot[fill=vizF!18, draw=vizF] coordinates {(165.5,igdn2F)};
      \addplot[fill=white,   draw=vizF, line width=1.1pt] coordinates {( 47.3,igdn2M)};
    \end{axis}
  \end{tikzpicture}

%% file: Sections/05RelatedWork.tex
\section{Related Work}
\paragraph{Linear RNNs.}
Linear RNNs, including state space models and linear attention, combine parallel training and efficient recurrent inference~\citep{martin2018parallelizing,katharopoulos2020transformers,dao2024transformers} with strong sequence modeling performance~\citep{gu2022efficiently,orvieto2023resurrecting,gu2023mamba}, making them promising alternatives to Transformers.
Representative models include S4~\citep{gu2022efficiently}, the Linear Recurrent Unit (LRU)~\citep{orvieto2023resurrecting}, RWKV~\citep{peng2023rwkv}, Mamba~\citep{gu2023mamba}, Mamba-2~\citep{dao2024transformers}, and Mamba-3~\citep{lahoti2026mamba}.
They also show promise on several tasks, including in-context learning~\citep{park2024mamba,grazzi2024mamba,li2025gating,jiang2025trained,oh2025mamba,bondaschi2026markov,li2026outliers}.
However, constrained state transitions and fixed-size memory can hinder state tracking, associative recall, and copying~\citep{merrill2024illusion,grazzi2025unlocking,arora2023zoology,jelassi2024repeat}.

\paragraph{Linear Attention.}
Linear attention~\citep{katharopoulos2020transformers} additively accumulates key--value associations, which can interfere with one another~\citep{schlag2021linear}.
To address this interference, RetNet~\citep{sun2023retentive} uses constant scalar decay and Mamba-2~\citep{dao2024transformers} and Gated Linear Attention (GLA)~\citep{yang2024gated} extend this mechanism to time-varying scalar and channel-wise decay, respectively.
DeltaNet~\citep{schlag2021linear,yang2024parallelizing} addresses interference through delta-rule updates for targeted overwriting of key--value memory, with later variants adding gating~\citep{yang2025gated,team2025kimi,hatamizadeh2026gated}.
The DeltaNet state update corresponds to one stochastic gradient descent (SGD) step on squared prediction error for values given keys.
This perspective motivates variants of DeltaNet that perform multiple updates per token~\citep{siems2026deltaproduct}, apply diagonal preconditioning~\citep{tumma2026preconditioned}, or incorporate momentum~\citep{huang2026mdn}. \citet{park2026qdelta} incorporate queries into state updates by combining key- and query-based prediction errors.

\paragraph{Neural Memory.}
From an online optimization view, DeltaNet can be interpreted as learning key--value associations by updating a linear model to minimize squared prediction error.
\citet{sun2025learning} extend this view by treating the parameters of a neural network, such as an MLP, as the recurrent state, with each state update corresponding to a gradient descent step on these parameters.
For these neural memory updates, \citet{behrouz2026titans} add momentum and \citet{behrouz2025atlas} use Muon~\citep{jordan2024muon}.
These memories offer greater expressiveness and can improve long-context modeling with memory size independent of sequence length.
However, nonlinear updates can increase computation and memory I/O while limiting parallelization relative to linear attention~\citep{sun2025learning,behrouz2026connected}.

\paragraph{Complementary Memories.}
Similar to our work, other approaches have explored combining complementary memories.
\citet{behrouz2026titans} combine local attention with neural memory and \citet{liu2026livemem} add a Gated DeltaNet-2 branch to a pretrained Transformer to retain information beyond its bounded attention window.
\citet{du2026mom} route tokens to independent recurrent states and \citet{bayat2026proteus} progressively activate memory as context grows.
Integral DeltaNet instead preserves information weakened by later updates through a weighted history of delta states.

\paragraph{Iterate Averaging.}
We form the integral state as an exponential moving average of delta states across tokens, starting from a zero initial state.
Averaging iterates across steps is also used in Stochastic Weight Averaging~\citep{izmailov2018averaging} and Lookahead~\citep{zhang2019lookahead}.
Other examples include schedule-free methods and their variants~\citep{defazio2024road,defazio2026schedulefree,kim2026amuse,apte2026anytime}: they combine a fast iterate with its running average and compute gradients at their interpolation, avoiding a prescribed learning-rate decay schedule.
\citet{song2025river} show that implicit weight averaging helps schedule-free methods track low-loss trajectories.
Despite this similarity, Integral DeltaNet differs from applying a schedule-free optimizer: we use the integral state for adaptive readout, not to update the delta state.

%% file: Sections/06Conclusion.tex
\section{Conclusion}\label{sec:conclusion}
In this work, we introduced Integral DeltaNet to address the trade-off between mitigating memory collisions and retaining useful information.
To this end, the model pairs a delta state that adapts to incoming key--value associations with an integral state that preserves a weighted history of delta states.
Adaptive readout combines these complementary states to access information weakened by later updates, while our chunkwise parallel formulation enables hardware-efficient training.
Experiments with Integral KDA and Integral GDN-2 show overall gains in language modeling, commonsense reasoning, retrieval, long-context understanding, and length extrapolation.
Our evaluation also compares training throughput and activation memory, as well as decoding latency and memory.
These results highlight the importance of integrated memory and adaptive readout.
A possible direction for future work is to use the integral state to guide delta-state updates, similar to how schedule-free methods~\citep{defazio2024road} use averaged iterates.

%% file: Sections/AProof.tex
\section{Proofs for the Illustrative Example}\label{app:event-retrieval}

\subsection{Proof of Proposition~\ref{prop:event-gdn}}\label{app:event-gdn}

Note that $\vk_1^\top\vk_2=\vk_1^\top\vk_3=1/2$ and $\vk_2^\top\vk_3=-1/2$.
For each final state $\mS_3$, define the Yes--No score differences
\[
 m_i:=(\vv_{\mathrm Y}-\vv_{\mathrm N})^\top\mS_3\vk_i,\qquad i=1,2,3.
\]
A positive score difference predicts Yes, and a negative one predicts No.
For linear attention, summing the three writes gives
\[
 \mS_3=\vv_{\mathrm N}\vk_1^\top+\vv_{\mathrm Y}(\vk_2+\vk_3)^\top
 =(\vv_{\mathrm N}+\vv_{\mathrm Y})\vk_1^\top.
\]
The answer-vector norms and orthogonality give $(m_1,m_2,m_3)=(-1,-1/2,-1/2)$. All three predictions are No, with risk $\mathcal R_{\mathrm{Linear\ Attention}}=2/3$.

For Mamba-2, applying the update $\mS_t=\alpha_t\mS_{t-1}+\vv_t\vk_t^\top$ three times gives
\[
 \mS_3=\alpha_2\alpha_3\vv_{\mathrm N}\vk_1^\top+\alpha_3\vv_{\mathrm Y}\vk_2^\top+\vv_{\mathrm Y}\vk_3^\top.
\]
With $(\alpha_1,\alpha_2,\alpha_3)=(1,1/4,1)$, the final state is $\mS_3=(\tfrac14\vv_{\mathrm N}+\vv_{\mathrm Y})\vk_1^\top$. The corresponding score differences are
\[
 (m_1,m_2,m_3)=\left(\tfrac12,\tfrac14,\tfrac14\right).
\]
All three predictions are Yes. Only the first query is incorrect, and $\mathcal R_{\mathrm{Mamba\textup{-}2}}=1/3$.

For DeltaNet, $\alpha_t=1$ in~\eqref{eq:event-gdn}. With $(\beta_1,\beta_2,\beta_3)=(1/4,1,1)$, the first two states are
\[
 \mS_1=\tfrac14\vv_{\mathrm N}\vk_1^\top,\qquad
 \mS_2=\tfrac14\vv_{\mathrm N}(\vk_1-\tfrac12\vk_2)^\top+\vv_{\mathrm Y}\vk_2^\top.
\]
These states satisfy $\mS_2\vk_2=\vv_{\mathrm Y}$ and $\mS_2\vk_3=\tfrac3{16}\vv_{\mathrm N}-\tfrac12\vv_{\mathrm Y}$. After the third write,
\begin{align*}
 \mS_3\vk_2
 &=\mS_2(\vk_2+\tfrac12\vk_3)-\tfrac12\vv_{\mathrm Y}
 =\tfrac3{32}\vv_{\mathrm N}+\tfrac14\vv_{\mathrm Y},\\
 \mS_3\vk_3&=\vv_{\mathrm Y},\qquad
 \mS_3\vk_1=\tfrac3{32}\vv_{\mathrm N}+\tfrac54\vv_{\mathrm Y}.
\end{align*}
The final score differences are
\[
 (m_1,m_2,m_3)=\left(\tfrac{17}{16},\tfrac1{16},1\right).
\]
Only the first query is incorrect, and $\mathcal R_{\mathrm{DeltaNet}}=1/3$.

Gated DeltaNet combines Mamba-2's decay with the delta rule and includes DeltaNet as the case $\alpha_t=1$. It therefore also attains risk $1/3$ with the DeltaNet gates above.

To show that Mamba-2, DeltaNet, and Gated DeltaNet cannot achieve risk below $1/3$, consider any final state $\mS_3$. Since $\vk_1=\vk_2+\vk_3$, the score differences satisfy $m_1=m_2+m_3$. Correctly answering the second and third queries requires $m_2,m_3>0$. This forces $m_1>0$, giving an incorrect Yes answer to the first query. Thus at least one of the three equally likely queries must be incorrect. Together with the constructions above, this proves that all three models have minimum risk $1/3$.
\hfill $\square$

\subsection{Proof of Proposition~\ref{prop:event-integral}}\label{app:event-integral}
Consider Gated DeltaNet with $\beta_t=1$ and $(\alpha_1,\alpha_2,\alpha_3)=(1,1/4,1)$. The three states are
\[
\begin{aligned}
 \mS_1 &= \vv_{\mathrm N}\vk_1^\top,\\
 \mS_2 &= \tfrac14\vv_{\mathrm N}(\vk_1-\tfrac12\vk_2)^\top+\vv_{\mathrm Y}\vk_2^\top,\\
 \mS_3 &= \tfrac14\vv_{\mathrm N}(\vk_1-\tfrac12\vk_2-\tfrac34\vk_3)^\top
          +\vv_{\mathrm Y}(\vk_2+\tfrac32\vk_3)^\top.
\end{aligned}
\]
The Yes--No score differences for the first query at the three write times are
\[
 (\vv_{\mathrm Y}-\vv_{\mathrm N})^\top\mS_1\vk_1=-2,\qquad
 (\vv_{\mathrm Y}-\vv_{\mathrm N})^\top\mS_2\vk_1=\tfrac18,\qquad
 (\vv_{\mathrm Y}-\vv_{\mathrm N})^\top\mS_3\vk_1=\tfrac{17}{16}.
\]
With $\lambda_t=1/8$ for $t=1,2,3$, applying $\mX_t=\lambda_t\mS_t+(1-\lambda_t)\mX_{t-1}$ three times from $\mX_0=\mzero$ gives
\[
 \begin{aligned}
 \mX_3 &= \lambda_1(1-\lambda_2)(1-\lambda_3)\mS_1+\lambda_2(1-\lambda_3)\mS_2+\lambda_3\mS_3\\
 &=\tfrac18\left(\tfrac{49}{64}\mS_1+\tfrac78\mS_2+\mS_3\right).
 \end{aligned}
\]
The score difference for the first query is
\[
 (\vv_{\mathrm Y}-\vv_{\mathrm N})^\top\mX_3\vk_1
 =\tfrac18\left(-\tfrac{49}{64}\cdot2+\tfrac78\cdot\tfrac18+\tfrac{17}{16}\right)
 =-\tfrac{23}{512}<0.
\]
Read $\mX_3$ for $\vq_3=\vk_1$ and $\mS_3$ for $\vq_3\in\{\vk_2,\vk_3\}$. For the latter two queries,
\[
 (\vv_{\mathrm Y}-\vv_{\mathrm N})^\top\mS_3\vk_2=\tfrac1{16}>0,\qquad
 (\vv_{\mathrm Y}-\vv_{\mathrm N})^\top\mS_3\vk_3=1>0.
\]
The Yes--No score difference is negative for the first query and positive for the other two. These signs give the correct answers No, Yes, and Yes, respectively, proving $\mathcal R_{\mathrm{Integral}}=0$.
\hfill $\square$

%% file: Sections/BGateAblation.tex
\section{Ablation Studies and Additional Analyses}\label{app:ablations}
We conduct additional experiments to validate three key design choices of Integral DeltaNet: (i) an integral state that accumulates the history of the delta state, (ii) an input-dependent readout that adaptively combines the two states, and (iii) channel-wise readout and integration gates.

\subsection{Effect of the Integral State}\label{app:integral-state}

To validate the effect of the integral state, we compare Integral KDA with models that share the KDA backbone but have larger recurrent states, using the 340M parameter, 15B token recipe in Section~\ref{sec:exp}.
KDA $d_k\times2$ and KDA $d_v\times2$ double the key and value dimensions, respectively.
To show that maintaining states at different timescales matters, rather than simply doubling the state, we also train a variant with two KDA states, which have independent decay and erase/write gates, share queries, keys, and values, and are combined with the same readout as Integral KDA.
For all variants, we adjust the FFN width to match the total parameter count.
Table~\ref{tab:integral-state} shows the performance of these models.
Integral KDA performs best overall, suggesting that the gains come from integrating the delta-state history at a slower timescale rather than from a larger state.

\begingroup
\begin{table}[htbp]
  \centering
  \small
  \setlength{\tabcolsep}{3pt}
  \renewcommand{\arraystretch}{1.12}
  \caption{Effect of the integral state (340M parameters, 15B tokens).}
  \label{tab:integral-state}
  \resizebox{\textwidth}{!}{%
  \begin{tabular}{@{}l@{\hspace{1em}}cc@{\hspace{0.8em}}ccccccccc@{\hspace{0.5em}}c@{\hspace{1.2em}}cccccc@{\hspace{0.5em}}c@{}}
    \toprule
    & \multicolumn{12}{c}{\textbf{Language modeling and commonsense reasoning}}
      & \multicolumn{7}{c}{\textbf{Real-world retrieval}} \\
    \cmidrule(lr){2-13}\cmidrule(l){14-20}
    Model & Wiki. & LMB. & LMB. & PIQA & Hella. & Wino. & ARC-e & ARC-c & OBQA & SIQA & BoolQ & Avg.
      & SWDE & SQuAD & FDA & TriviaQA & NQ & DROP & Avg. \\
    & ppl $\downarrow$ & ppl $\downarrow$ & acc $\uparrow$ & acc $\uparrow$ & acc$_n$ $\uparrow$
      & acc $\uparrow$ & acc $\uparrow$ & acc$_n$ $\uparrow$ & acc$_n$ $\uparrow$ & acc $\uparrow$
      & acc $\uparrow$ & acc $\uparrow$ & \multicolumn{7}{c}{acc $\uparrow$} \\
    \midrule
    KDA & \textbf{24.89} & 26.16 & 36.4 & 66.9 & 41.2 & 53.4 & 59.4 & 27.3 & 33.6 & 39.0 & 57.9 & 46.1 & 10.2 & 28.6 & 8.1 & 47.8 & 14.4 & 17.7 & 21.1 \\
    KDA $d_k\times2$ & 25.87 & 25.88 & 35.8 & 66.8 & 40.5 & \textbf{54.3} & 58.5 & 27.0 & 32.0 & 38.7 & 60.9 & 46.1 & 9.3 & 30.2 & 7.6 & 48.4 & 14.6 & 16.4 & 21.1 \\
    KDA $d_v\times2$ & 24.95 & \textbf{24.40} & \textbf{37.4} & 66.7 & \textbf{41.3} & 53.3 & \textbf{61.7} & 27.5 & 33.6 & 38.7 & 50.3 & 45.6 & 15.1 & 30.5 & 7.6 & 46.8 & 14.0 & \textbf{19.1} & 22.2 \\
    Two KDA states & 25.10 & 25.40 & 36.3 & 65.9 & 40.8 & 52.0 & 57.9 & 27.1 & 34.6 & \textbf{39.4} & \textbf{61.7} & 46.2 & 13.0 & 29.7 & 7.6 & 48.6 & 14.9 & 18.2 & 22.0 \\
    \midrule
    Integral KDA & 24.97 & 24.99 & 36.9 & \textbf{68.4} & 40.7 & 52.9 & 58.2 & \textbf{28.0} & \textbf{34.8} & 38.8 & 59.5 & \textbf{46.5} & \textbf{18.1} & \textbf{30.6} & \textbf{9.6} & \textbf{48.9} & \textbf{15.0} & 17.3 & \textbf{23.3} \\
    \bottomrule
  \end{tabular}%
  }
\end{table}
\endgroup

\subsection{Effect of Input-Dependent Readout}\label{app:readout}

We examine the input-dependent readout in two ways: an ablation that removes its input dependence at the 340M scale, and an analysis of the learned readout in the 1.3B model from the main experiments.

\paragraph{Constant Readout Gate.}
To validate the effect of input-dependent readout, we replace the readout gate of Integral KDA with a constant $\vgamma_t=\bar{\vgamma}$, learned per channel and shared across tokens, while keeping the input-dependent integration gate.
We train this variant with the 340M parameter, 15B token recipe in Section~\ref{sec:exp}.
Table~\ref{tab:readout-ablation} shows the performance of these models.
The constant readout gate retains a modest retrieval gain over KDA but lowers both averages relative to Integral KDA, suggesting that adapting the readout to the input is beneficial.

\begingroup
\begin{table}[H]
  \centering
  \small
  \setlength{\tabcolsep}{3pt}
  \renewcommand{\arraystretch}{1.12}
  \caption{Effect of input-dependent readout (340M parameters, 15B tokens).}
  \label{tab:readout-ablation}
  \resizebox{\textwidth}{!}{%
  \begin{tabular}{@{}l@{\hspace{1em}}cc@{\hspace{0.8em}}ccccccccc@{\hspace{0.5em}}c@{\hspace{1.2em}}cccccc@{\hspace{0.5em}}c@{}}
    \toprule
    & \multicolumn{12}{c}{\textbf{Language modeling and commonsense reasoning}}
      & \multicolumn{7}{c}{\textbf{Real-world retrieval}} \\
    \cmidrule(lr){2-13}\cmidrule(l){14-20}
    Model & Wiki. & LMB. & LMB. & PIQA & Hella. & Wino. & ARC-e & ARC-c & OBQA & SIQA & BoolQ & Avg.
      & SWDE & SQuAD & FDA & TriviaQA & NQ & DROP & Avg. \\
    & ppl $\downarrow$ & ppl $\downarrow$ & acc $\uparrow$ & acc $\uparrow$ & acc$_n$ $\uparrow$
      & acc $\uparrow$ & acc $\uparrow$ & acc$_n$ $\uparrow$ & acc$_n$ $\uparrow$ & acc $\uparrow$
      & acc $\uparrow$ & acc $\uparrow$ & \multicolumn{7}{c}{acc $\uparrow$} \\
    \midrule
    KDA & \textbf{24.89} & 26.16 & 36.4 & 66.9 & \textbf{41.2} & \textbf{53.4} & \textbf{59.4} & 27.3 & 33.6 & \textbf{39.0} & 57.9 & 46.1 & 10.2 & 28.6 & 8.1 & 47.8 & 14.4 & 17.7 & 21.1 \\
    Constant readout gate & 25.26 & 25.04 & 36.7 & 67.1 & 40.7 & 50.5 & 59.0 & 27.7 & 33.8 & \textbf{39.0} & 58.5 & 45.9 & 12.7 & 28.9 & \textbf{9.6} & 46.9 & 13.2 & \textbf{19.5} & 21.8 \\
    \midrule
    Integral KDA & 24.97 & \textbf{24.99} & \textbf{36.9} & \textbf{68.4} & 40.7 & 52.9 & 58.2 & \textbf{28.0} & \textbf{34.8} & 38.8 & \textbf{59.5} & \textbf{46.5} & \textbf{18.1} & \textbf{30.6} & \textbf{9.6} & \textbf{48.9} & \textbf{15.0} & 17.3 & \textbf{23.3} \\
    \bottomrule
  \end{tabular}%
  }
\end{table}
\endgroup

\paragraph{Learned Readout Behavior.}\label{app:readout-behavior}

We next analyze how the learned readout uses the two states in the 1.3B Integral KDA trained on 100B tokens, which is the same model evaluated in Section~\ref{sec:exp}.
On S-NIAH-2 (4K), we compare \emph{distant needles} ($3{,}122$--$3{,}922$ tokens before the answer) and \emph{recent needles} ($44$--$132$ tokens).
For each sample and head in each layer, we compute the following ratio at the answer position $t$:
\begingroup
\setlength{\abovedisplayshortskip}{2pt}
\setlength{\belowdisplayshortskip}{4pt}
\[
  \text{Integral-state contribution}
  = \frac{\lVert\mX_t((\vone-\vgamma_t)\odot\vq_t)\rVert_2}
         {\lVert\mS_t(\vgamma_t\odot\vq_t)\rVert_2+\lVert\mX_t((\vone-\vgamma_t)\odot\vq_t)\rVert_2}.
\]
\endgroup
Figure~\ref{fig:contrib-depth} plots this ratio by layer, pooling one value per sample/head pair within each needle group. While some layers show little separation, layers 15 and 16 show markedly greater integral contributions for distant than recent needles, consistent with using accumulated memory for distant retrieval.

\begingroup
\begin{figure}[!htb]

  \input{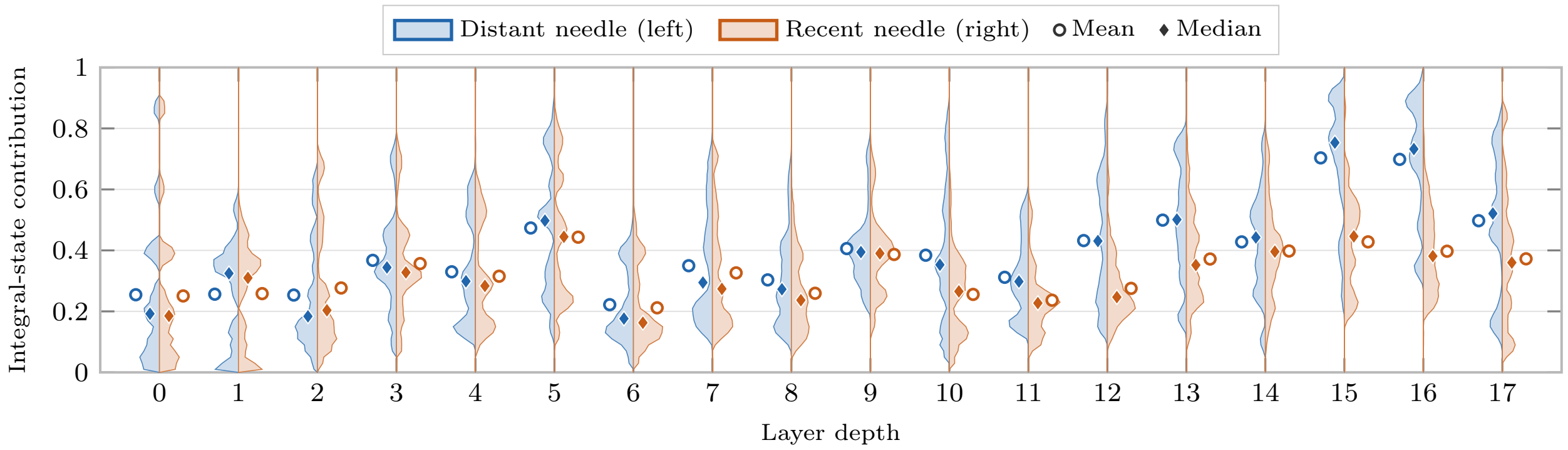}
  \caption{Integral-state contribution by layer. Plots show distributions over sample/head pairs for distant (left) and recent (right) needles; open circles and filled diamonds mark means and medians.}
  \label{fig:contrib-depth}
\par

\end{figure}
\endgroup

\subsection{Effect of Channel-Wise Gating}\label{app:channel-gating}

Integral DeltaNet uses channel-wise gates so that each key channel has its own readout weight and integration rate, allowing different channels to operate at different timescales.
To validate the effect of channel-wise gating, we compare all four combinations of scalar and channel-wise readout ($\gamma$) and integration ($\lambda$) gates in Integral KDA, using the 340M parameter, 15B token recipe in Section~\ref{sec:exp}.
All variants share the same initial gate values.
Table~\ref{tab:kda-gate-ablation} shows the performance of these variants.
Using channel-wise gates for both the readout and integration achieves the lowest perplexities and the highest commonsense average, suggesting that allowing each key channel to adjust its own readout weight and integration rate is beneficial.

\begingroup
\begin{table}[htbp]

  \centering
  \small
  \setlength{\tabcolsep}{3pt}
  \renewcommand{\arraystretch}{1.12}
  \caption{Effect of channel-wise gating (340M parameters, 15B tokens). S: scalar; C: channel-wise.}
  \label{tab:kda-gate-ablation}
  \resizebox{\textwidth}{!}{%
  \begin{tabular}{@{}cc@{\hspace{1em}}cc@{\hspace{0.8em}}ccccccccc@{\hspace{0.5em}}c@{\hspace{1.2em}}cccccc@{\hspace{0.5em}}c@{}}
    \toprule
    & & \multicolumn{12}{c}{\textbf{Language modeling and commonsense reasoning}}
      & \multicolumn{7}{c}{\textbf{Real-world retrieval}} \\
    \cmidrule(lr){3-14}\cmidrule(l){15-21}
    $\gamma$ & $\lambda$ & Wiki. & LMB. & LMB. & PIQA & Hella. & Wino. & ARC-e & ARC-c & OBQA & SIQA & BoolQ & Avg.
      & SWDE & SQuAD & FDA & TriviaQA & NQ & DROP & Avg. \\
    & & ppl $\downarrow$ & ppl $\downarrow$ & acc $\uparrow$ & acc $\uparrow$ & acc$_n$ $\uparrow$
      & acc $\uparrow$ & acc $\uparrow$ & acc$_n$ $\uparrow$ & acc$_n$ $\uparrow$ & acc $\uparrow$
      & acc $\uparrow$ & acc $\uparrow$ & \multicolumn{7}{c}{acc $\uparrow$} \\
    \midrule
    S & S & 25.18 & 28.31 & 34.1 & 67.2 & 41.1 & 51.8 & 57.4 & 27.3 & 33.6 & \textbf{39.6} & 53.5 & 45.1 & 14.0 & 29.0 & 9.1 & 47.4 & 14.6 & 18.9 & 22.2 \\
    C & S & 25.08 & 25.30 & \textbf{37.0} & 67.0 & \textbf{41.2} & 50.6 & \textbf{59.8} & \textbf{28.2} & 32.0 & 38.2 & \textbf{59.7} & 46.0 & 16.2 & 30.0 & 11.5 & 47.6 & 14.6 & 18.3 & 23.0 \\
    S & C & 25.07 & 26.06 & 36.1 & 66.3 & 41.0 & 51.3 & 58.6 & 28.0 & \textbf{34.8} & 39.0 & 57.5 & 45.8 & 16.6 & 29.1 & \textbf{13.4} & 48.0 & 14.9 & \textbf{20.7} & \textbf{23.8} \\
    C & C & \textbf{24.97} & \textbf{24.99} & 36.9 & \textbf{68.4} & 40.7 & \textbf{52.9} & 58.2 & 28.0 & \textbf{34.8} & 38.8 & 59.5 & \textbf{46.5} & \textbf{18.1} & \textbf{30.6} & 9.6 & \textbf{48.9} & \textbf{15.0} & 17.3 & 23.3 \\
    \bottomrule
  \end{tabular}%
  }
\par

\end{table}
\endgroup